\documentclass{article} 
\usepackage{iclr2027_conference,times}
\usepackage{fontawesome5}

\usepackage{amsmath,amsfonts,bm}

\def\eqref#1{equation~\ref{#1}}

\def\1{\bm{1}}

\DeclareMathAlphabet{\mathsfit}{\encodingdefault}{\sfdefault}{m}{sl}
\SetMathAlphabet{\mathsfit}{bold}{\encodingdefault}{\sfdefault}{bx}{n}

\usepackage{multirow}
\usepackage[
    pagebackref,
    breaklinks,
    citecolor=uclablue,
    linkcolor=uclablue,
    colorlinks=true,
]{hyperref}
\definecolor{uclablue}{rgb}{0.15, 0.45, 0.68}
\usepackage{algorithm}
\usepackage{algpseudocode}
\usepackage{url}
\usepackage{booktabs}
\usepackage{graphicx}
\usepackage{cleveref}
\newcommand{\model}{\textsc{T2-z}}
\usepackage{xcolor}
\newcommand{\algcomment}[1]{\textcolor{gray}{\textit{// #1}}}

\title{\model: Test-Time Optimization of Environment Latents for Robotic World Models}

\author{\parbox{0.97\textwidth}{\centering\normalfont
Chenyuan Zhou\textsuperscript{1}*, Yining Hong\textsuperscript{1}*,
Huang Huang\textsuperscript{1}$^\heartsuit$,
Yunfan Jiang\textsuperscript{1}$^\heartsuit$,\\[0.2em]
Wenlong Huang\textsuperscript{1}$^\heartsuit$,
David Held\textsuperscript{2},
Vitor Campagnolo Guizilini\textsuperscript{3},
Yue Wang\textsuperscript{4},\\[0.2em]
Li Fei-Fei\textsuperscript{1},
Jiajun Wu\textsuperscript{1}$^\dagger$,
Leonidas Guibas\textsuperscript{1}$^\dagger$,
Yejin Choi\textsuperscript{1}$^\dagger$\\[0.6em]
{\footnotesize
* Co-First Contribution\quad
$\heartsuit$ Co-Second Contribution \quad
$\dagger$ Equal Advising \\[1em]
{\small
\textsuperscript{1}Stanford University\quad
\textsuperscript{2}CMU\\
\textsuperscript{3}Toyota Research Institute\quad
\textsuperscript{4}University of Southern California}}\\[1em]
{\small
\faGlobe\ \textbf{Website:} \url{https://t2-z-latent-optimization.github.io/}\\[0.2em]
\faGithub\ \textbf{Code:} \url{https://github.com/T2-z-latent-optimization/T2-z}\\[0.2em]
\faDatabase\ \textbf{Data:} \url{https://huggingface.co/datasets/huagailuowen/T2-z-datasets}
}
}}

\iclrfinalcopy 
\begin{document}

\maketitle
\begin{figure*}[htbp]
\vspace{-1.5em}
    \centering
    \includegraphics[width=0.95\linewidth]{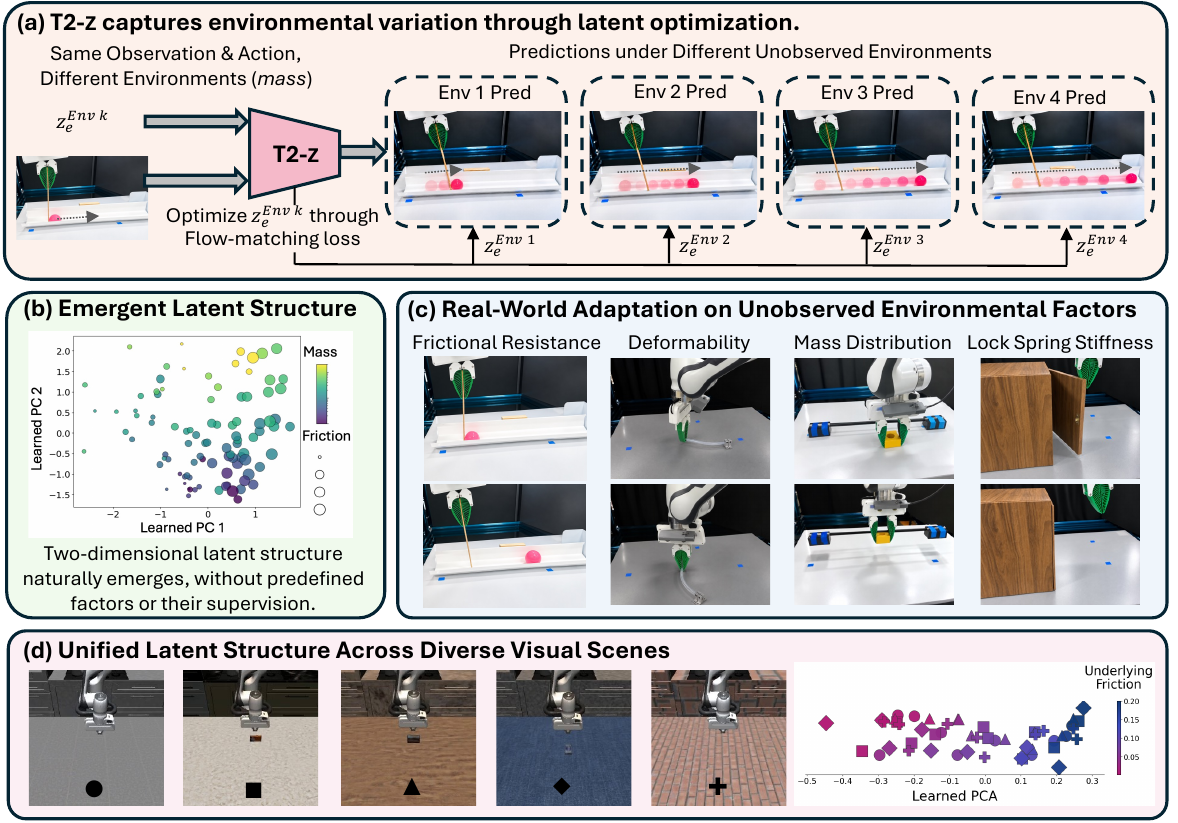}
    \vspace{-0.7em}
    \caption{\small \textbf{\model{} for Test-Time Environment Latent Optimization.}
\textbf{(a)} Given the same observation and action, \model{} directly optimizes
environment latents to capture different environmental conditions
(\textit{e.g.,} ball mass), leading the world model to predict different outcomes. 
\textbf{(b)} Without knowing how many factors exist or what they represent, PCA of the learned
environment latents reveals a two-dimensional structure that closely reflects
the underlying factor distribution.
\textbf{(c)} We evaluate \model{} on real-world robotic tasks spanning diverse
unobserved environmental factors.
\textbf{(d)} Across diverse visual scenes, the learned
latents reveal a unified structure that reflects the same underlying
environmental factor without being entangled with visual appearance. 
}
    \label{fig:teaser}
\end{figure*}
\begin{abstract}
\vspace{-0.8em}
Predicting how the world responds to an action requires more than what can be seen: robots must account for environmental factors that are neither directly observable nor explicitly specified, from physical properties such as friction and mass to human-defined environmental rules. 
We introduce \model{} for test-time optimization of environment latents, which augments a robotic world model with a learnable environment latent to condition world predictions. This compact latent captures underlying environmental variation without supervision of what environmental factors exist or what values they take.
During training, \model{} alternates between two stages: a world-model optimization stage that fixes the latents and learns general action-conditioned dynamics across environments; and an environment-latent optimization stage that freezes the world model and optimizes the conditioning latents, allowing latent structure to emerge purely through world prediction.
At test time, the world model remains frozen and a newly initialized environment latent is optimized from one or more test-time interactions, adapting the model to previously unseen environmental configurations. 
Across 7 simulated settings and 4 real-world robotic tasks, \model{} reduces object prediction error metric from $15.20$ to $7.00$ and raises action success from $33.73\%$ to $72.96\%$.
Its emergent latents capture structured
representations of individual and compounded factors across diverse visual scenes.
Ablations further show that \model{} can efficiently infer environment latents from as little as a single test-time interaction trajectory. \looseness=-1


\end{abstract}

\vspace{-1em}
\section{Introduction}
\vspace{-0.5em}
What a robot observes is insufficient to determine how the world will respond. Beneath visually similar scene configurations reside hidden environmental factors, such as friction, mass distribution, and unobserved human-defined interaction rules, that can fundamentally alter the consequences of the same action (\cref{fig:teaser} (a)). Such factors are often revealed only through interaction with the target environments, as their influence becomes observable through the outcomes they induce. Explicitly representing these factors \citep{kumar2021rma}, however, requires specifying in advance what factors may vary, their exact values, and how they should be parameterized, which is difficult to scale to unknown or compounded environmental variation. On the contrary, humans reason through ``analysis by synthesis" ~\citep{yuille2006vision}, using internal generative models not only to anticipate future outcomes, but also in reverse to infer latent structure that best accounts for what they observe. This raises a natural question: can hidden environmental factors be similarly captured in a latent representation that is directly optimized from test-time interactions?

We introduce \model{} for test-time optimization of environmental latents, which augments an action-conditioned robotic world model with a learnable environment latent $\mathbf{z}_e$. $\mathbf{z}_e$ is a randomly-initialized compact representation that is expected to encode the unobserved factors governing how an environment responds to interaction, with no supervision about what environmental factors it should represent. 
The world model predicts future visual observations conditioned jointly on the action and $\mathbf{z}_e$. We iteratively optimize the world model and $\mathbf{z}_e$ in an alternating two-stage procedure: 1) world-model optimization: with $\mathbf{z}_e$ fixed, we train the world model across all trajectories to learn shared action-conditioned dynamics across environments; 2) environment-latent optimization: with the world model frozen, we optimize the conditioned latent $\mathbf{z}_e$ through prediction error, so that it captures underlying environment-specific variation not explained by the general world model alone. Both stages are optimized with the same flow-matching loss and alternated for multiple rounds until convergence, allowing the environment representation to emerge directly through prediction without predefined physical-factor supervision or a separate environment encoder.

At test time, the world model is frozen and \model{} turns environment inference into a direct
optimization problem. For a previously unseen environmental configuration,
we initialize a new $\mathbf{z}_e$ and optimize $\mathbf{z}_e$ to explain the outcomes of one or more observed test-time interactions.
Specifically, we update $\mathbf{z}_e$ for multiple optimization steps through the same flow-matching objective used
during stage-two training, minimizing prediction error on the observed interactions. Because inference is optimization-based rather than amortized through a learned history encoder, \model{} is not tied to the trajectory count used for training and can flexibly incorporate any number of test-time trajectories.
Once optimized, the latent turns the frozen world model into an
environment-adapted test-time simulator, enabling prediction and comparison
of candidate action outcomes before execution.


We evaluate \model{} across 7 simulated settings and 4 real-world robotic tasks. Quantitatively and qualitatively, \model{} improves world-model prediction across visual quality, physical prediction metrics (\textit{e.g.}, relevant object locations), and downstream action selection. \model{} reduces physical prediction error from $12.86$ to $4.80$ in simulation and tracked-state error from $19.30$ to $10.85$ pixels in the real world, while improving action success from $37.9\%$ to $74.2\%$ and from $26.5\%$ to $70.8\%$, respectively. PCA of the learned latents reveals structured representations that reflect the underlying environmental factors, including a multi-dimensional organization when more than one factors vary jointly, despite the model never knowing how many factors are present or what they represent (\cref{fig:teaser}(b)). In our multi-background setting, a unified latent structure emerges across diverse scene setups, revealing that the learned representation captures environmental dynamics without becoming entangled with scene appearance (\cref{fig:teaser}(d)). Ablations further show that \model{} can infer useful environment representations from as few as a single test-time interaction trajectory.

\section{Related Works}
\begin{figure*}[htbp]
    \centering
    \includegraphics[width=\linewidth]{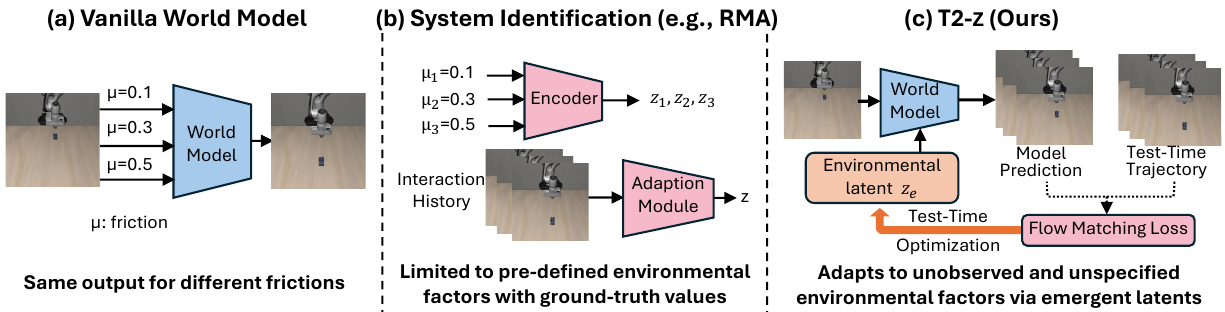}
    \caption{\small \textbf{Comparison with existing approaches.}
\textbf{(a)} Vanilla world models do not explicitly represent unobserved
environmental variation, producing the same prediction under different
environmental conditions.
\textbf{(b)} System-identification approaches such as RMA learn environment
representations from predefined factors and their ground-truth values, then
infer them from interaction history through a learned encoder.
\textbf{(c)} \model{} requires neither predefined environmental factors nor a
learned environment encoder, instead directly optimizing an emergent
environment latent from test-time interaction through prediction error.
}
    \label{fig:related}
\end{figure*}
Two distinct strands of work are particularly related to our problem.

\textbf{Robotic world models} have made rapid progress in predicting
action-conditioned futures~\citep{zhou2025dinowm,zhou2024robodreamer,
zhu2024irasim,zhen2025tesseract,zhu2025uwm,ye2026dreamzero,
huang2026pointworld,gao2026dreamdojo}. More recent work explores test-time
adaptation to improve long-horizon memory or incorporate new
behavioral supervision~\citep{wan2026worldagen,wang2026adajepa,
feng2026wamttt,yoo2025wormi,jang2026tmow}. These approaches predict future outcomes from visual observations alone, without accounting for underlying environmental variation that is not directly observable. (\cref{fig:related} (a)).
In contrast, \model{} 
disentangles unobserved environmental variation from the world model's
observable scene representation, through environment latent conditioning. \looseness=-1

\textbf{Environment representations and system identification} have
demonstrated the value of environment-specific representations for system
identification and rapid behavioral adaptation
~\citep{yu2017preparing,nagabandi2019adapt,rakelly2019pearl,lee2020cadm,
kumar2021rma,miki2022robust,nahrendra2023dreamwaq,wang2023protocad,
he2024hib,luo2024pie}. The Rapid Motor Adaptation (RMA) family and related
methods typically rely on predefined privileged parameters
~\citep{kumar2021rma,miki2022robust,nahrendra2023dreamwaq,he2024hib,luo2024pie}. For example, RMA restricts
adaptation to 5 factors and uses their ground-truth
 values to learn an embedding that an adaptation encoder later
recovers from interaction history~\citep{kumar2021rma}. This prespecifies the
environment representation and ties inference to a learned
history-to-embedding mapping, limiting flexibility under unknown, compounded,
or out-of-distribution variation (\cref{fig:related} (b)). These methods also generally rely on a learned history encoder~\citep{chen2024slr,xing2026vnd}, \textit{limiting generalization when the amount of test-time interaction differs from that seen during training}.  In contrast, \model{} learns representations of environmental factors
implicitly through world prediction error, without predefined ground-truth physical parameters or a learned
history encoder. 

Our method is also related to the literature
on \textbf{Test-Time Adaptation} (TTA), which
 has enabled robots
to improve after deployment by adapting policy, prompts, or memory rather than optimizing a dedicated
environment representation. Existing approaches include test-time adaptation of
diffusion and flow policies~\citep{li2025adpro,li2026prigo},
online policy adaptation~\citep{liu2026ttvla,jin2026agenticvla,
zhu2026tttparkour}, and adaptation through latent prompts, fast weights,
memory, or retrieval~\citep{zhang2026tttvla,jiang2026robottt,feng2026wamttt,
zhao2026retrievesteer,park2026retrieve}. TTA has also been
applied directly to predictive models through video replanning, world-model
training, and closed-loop latent adaptation~\citep{ko2025ise,wan2026worldagen,
wang2026adajepa}. \looseness=-1

\section{Method}
\begin{figure*}[htbp]
    \centering
    \includegraphics[width=\linewidth]{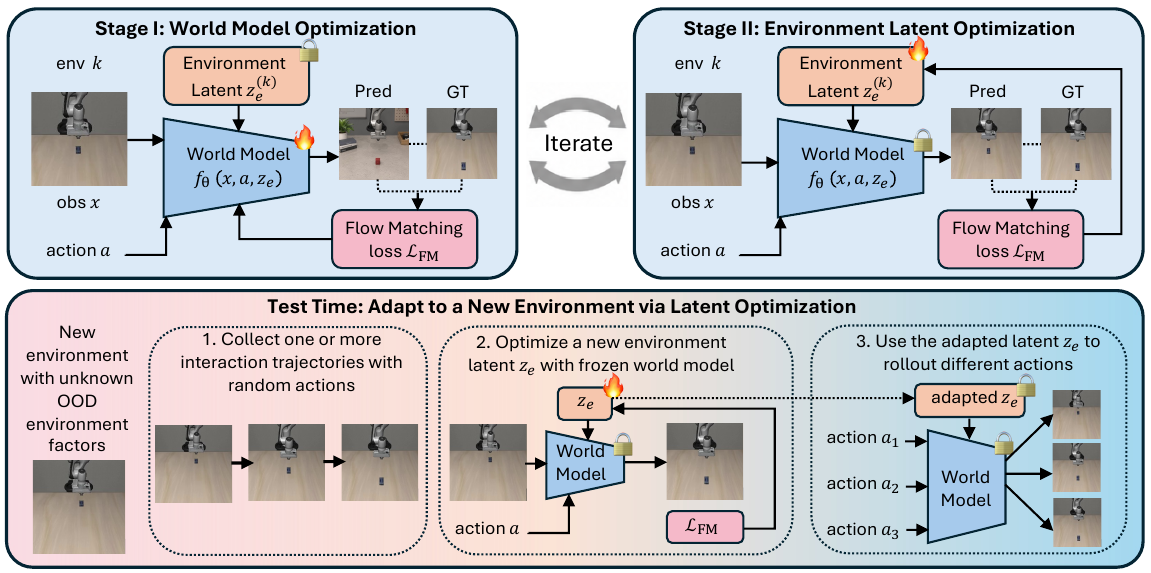}
    \caption{\small \textbf{\model{} framework.}
During training, \model{} alternates between two stages:
(1) world-model optimization, which learns shared action-conditioned
dynamics with environment latents fixed, and
(2) environment-latent optimization, which optimizes the latents
through prediction error with the world model frozen.
(3) At test time, the world model remains frozen while a new
environment latent is directly optimized from interaction and then reused
to predict the outcomes of different actions.
}
    \label{fig:method}
\end{figure*}

\label{sec:method}
We introduce \model{} (\cref{fig:related} \& \cref{alg:t2z}), a framework for adapting robotic world models to hidden environmental variation through direct optimization of an environment latent. Our key idea is to separate the general visual predictive capability of a world model from unobserved environment-specific information. During training, we learn these two components through alternating optimization under a shared flow-matching objective. At test time, we freeze the world model and optimize only a newly initialized
environment latent from observed test-time interaction trajectories. The optimized latent subsequently conditions the frozen world model to predict the outcomes of diverse actions.

\subsection{Problem Setup}
\label{sec:problem_setup}

We consider a robotic world model that predicts future visual observations $\mathbf{y}$ from the observation $\mathbf{x}$ and action $\mathbf{a}$. 
We augment the world model with an environment latent $\mathbf{z}_e$ and formulate prediction as:
\begin{equation}
    p_{\theta}
    \left(
        \mathbf{y}
        \mid
        \mathbf{x},
        \mathbf{a},
        \mathbf{z}_e
    \right),
\end{equation}
where $\theta$ denotes the shared world-model parameters and $\mathbf{z}_e$ is a latent representation of the environment.
Importantly, $\mathbf{z}_e$ receives no supervision of which environmental factors it should represent.

Let $\mathbf{y}_t$ denote the noisy/interpolated future observation
at flow timestep $t$, and $\mathbf{u}_t$ its corresponding target flow
velocity. The model is optimized using the standard flow-matching objective,
\begin{equation}
\label{eq:flow_matching}
    \mathcal{L}_{\mathrm{FM}}(\theta,\mathbf{z}_e)
    =
    \mathbb{E}
    \left[
        \left\|
        \mathbf{v}_{\theta}
        (\mathbf{y}_t,t
        \mid
        \mathbf{x},\mathbf{a},\mathbf{z}_e)
        -
        \mathbf{u}_t
        \right\|_2^2
    \right].
\end{equation}

\subsubsection{Training}
\label{sec:training}
For each training environment $e$, we maintain an environment latent
$\mathbf{z}_e$, randomly initialized by sampling each dimension from a
uniform distribution:
\begin{equation}
    \mathbf{z}_e^{(0)} \sim \mathcal{U}(0,1).
\end{equation}
 We alternate between two stages throughout training: world-model optimization and environment-latent optimization, with $k$ indexing the outer optimization cycles. The world model is updated with low learning rate to learn general dynamics
shared across environments, while the environment latents are optimized
rapidly for quick adaptation to environment information revealed
through interaction. This slow--fast separation follows the perspective of
learning at different timescales~\citep{tiwari2026learning} and is closely
related to test-time learning with slow/fast weights~\citep{hong2024slowfastvgenslowfastlearningactiondriven}. 

\begin{algorithm}[htbp]

\caption{\model{} Training and Test-Time Optimization}
\label{alg:t2z}
\footnotesize
\begin{minipage}[t]{0.49\linewidth}
\vspace{0pt}
\begin{algorithmic}[1]
\Require Training sets $\{\mathcal{D}_e\}_{e=1}^{E}$,
world model $p_\theta$, loss $\mathcal{L}_{\mathrm{FM}}$, learning rate $\eta_\theta$ and $\eta_z$
\State Initialize $\theta$ and
$\mathbf{z}_e \sim \mathcal{U}(0,1)$ for each environment $e$

\Statex
\Statex \algcomment{Training: alternating optimization}
\While{not converged}

    \Statex \algcomment{Stage I: world-model optimization}
    \State Freeze environment latents
    $\{\mathbf{z}_e\}_{e=1}^{E}$

    \For{world-model optimization steps}
        \State $\theta \gets \theta -
        \eta_\theta \nabla_\theta
        \mathcal{L}_{\mathrm{FM}}
        (\theta,\mathbf{z}_e)$
    \EndFor

    \Statex
    \Statex \algcomment{Stage II: environment-latent optimization}
    \State Freeze world-model parameters $\theta$

\algstore{t2zsplit}
\end{algorithmic}
\end{minipage}
\hfill
\begin{minipage}[t]{0.49\linewidth}
\vspace{0pt}
\begin{algorithmic}[1]
\algrestore{t2zsplit}

    \For{environment-latent optimization steps}
        \State $\mathbf{z}_e \gets \mathbf{z}_e -
        \eta_z \nabla_{\mathbf{z}_e}
        \mathcal{L}_{\mathrm{FM}}
        (\theta,\mathbf{z}_e)$
    \EndFor

\EndWhile

\Statex
\Statex \algcomment{Test-time optimization}
\Require Trained world model $\theta^*$,
observed interactions $\mathcal{D}_{\mathrm{test}}$, Optimization Step $s$, learning rate $\eta_z^{(s)}$
\State Freeze $\theta^*$ and initialize
$\mathbf{z}_{\mathrm{test}}^{(0)}$

\For{$s=0,\ldots,S-1$}
    \State $\mathcal{L}_{\mathrm{test}} \gets
    \frac{1}{|\mathcal{D}_{\mathrm{test}}|}
    \sum_{(\mathbf{x},\mathbf{a},\mathbf{y})
    \in\mathcal{D}_{\mathrm{test}}}
    \mathcal{L}_{\mathrm{FM}}
    (\theta^*,\mathbf{z}_{\mathrm{test}}^{(s)})$

    \State $\mathbf{z}_{\mathrm{test}}^{(s+1)}
    \gets \mathbf{z}_{\mathrm{test}}^{(s)}
    - \eta_z^{(s)}
    \nabla_{\mathbf{z}_{\mathrm{test}}^{(s)}}
    \mathcal{L}_{\mathrm{test}}$
\EndFor

\State \Return $\mathbf{z}_{\mathrm{test}}^*
\gets \mathbf{z}_{\mathrm{test}}^{(S)}$

\end{algorithmic}
\end{minipage}

\end{algorithm}

\paragraph{Stage I: World-Model Optimization.}
In the first stage, we fix all $\mathbf{z}_e$ and optimize only the world-model parameters. Let $\mathcal{D}_e$ denote the interaction trajectories collected under environment $e$.
\begin{equation}
\label{eq:world_model_optimization}
    \theta^{(k+1)}
    =
    \arg\min_{\theta}
    \mathbb{E}_{e,\,
    (\mathbf{x},\mathbf{a},\mathbf{y})\sim\mathcal{D}_e}
    \left[
    \mathcal{L}_{\mathrm{FM}}
    \left(
        \theta,
        \mathbf{z}_e^{(k)}
    \right)
    \right].
\end{equation}
During this stage, $\mathbf{z}_e$ is held constant, while the same world model is optimized over trajectories spanning different environments and actions. This encourages $\theta$ to capture predictive structure that is shared across the training distribution, including how visual states evolve to different robot actions. 

\paragraph{Stage II: Environment-Latent Optimization.}
In the second stage, we reverse the optimization: the world model is frozen, and only the environment latents are updated. For each environment $e$, \looseness=-1
\begin{equation}
\label{eq:environment_latent_optimization}
    \mathbf{z}_e^{(k+1)}
    =
    \arg\min_{\mathbf{z}_e}
    \mathbb{E}_{(\mathbf{x},\mathbf{a},\mathbf{y})\sim\mathcal{D}_e}
    \left[
    \mathcal{L}_{\mathrm{FM}}
    \left(
        \theta^{(k+1)},
        \mathbf{z}_e
    \right)
    \right].
\end{equation}
Because $\theta$ is fixed, the only way to reduce prediction error is to change $\mathbf{z}_e$ to better match the interaction outcomes observed in environment $e$.
As a result, environments that induce systematically different outcomes drive their latents toward different regions of the latent space.



\paragraph{Alternating Optimization}
Training starts with Stage I, then alternates between Stage II and Stage I, each optimized for a fixed number of steps while keeping the other component fixed. As the
world model improves, latents are optimized against an increasingly
capable predictive model; conversely, the refined latents provide
more informative conditioning for subsequent world-model optimization. \looseness=-1

\subsubsection{Test-Time Optimization}
\label{sec:test_time_latent}

At test time, we adapt to a previously unseen environment by optimizing only
its environment latent while keeping the trained world model fixed. We
initialize a new latent $\mathbf{z}_{\mathrm{test}}^{(0)}$ and observe one or
more interaction trajectories on randomly-sampled actions. We denote the resulting test-time interaction
set as 
\begin{equation}
    \mathcal{D}_{\mathrm{test}}
    =
    \left\{
    (\mathbf{x}_i,\mathbf{a}_i,\mathbf{y}_i)
    \right\}_{i=1}^{N},
\end{equation}
where each tuple consists of an observation $\mathbf{x}_i$, an
action $\mathbf{a}_i$, and the resulting observations
$\mathbf{y}_i$.

We then optimize $\mathbf{z}_{\mathrm{test}}$ using the same flow-matching
objective employed during training:
\begin{equation}
\label{eq:test_time_latent_optimization}
    \mathbf{z}_{\mathrm{test}}^{(s+1)}
    =
    \mathbf{z}_{\mathrm{test}}^{(s)}
    -
    \eta_z^{(s)}
    \nabla_{\mathbf{z}_{\mathrm{test}}^{(s)}}
    \frac{1}{|\mathcal{D}_{\mathrm{test}}|}
    \sum_{(\mathbf{x},\mathbf{a},\mathbf{y})\in\mathcal{D}_{\mathrm{test}}}
    \mathcal{L}_{\mathrm{FM}}
    \left(
        \theta^{*},
        \mathbf{z}_{\mathrm{test}}^{(s)};
        \mathbf{x},
        \mathbf{a},
        \mathbf{y}
    \right).
\end{equation}
where $\theta^{*}$ denotes the frozen world-model parameters and
$\eta_z^{(s)}$ follows a coarse-to-fine learning-rate strategy, with larger learning rates in early optimization steps and
smaller learning rates in later steps.
Because the world model is fixed, prediction errors over the observed
interactions can only be reduced by updating $\mathbf{z}_{\mathrm{test}}$.
The latent is therefore optimized toward a representation under which the
frozen world model reproduces the interaction outcomes observed in the
current environment.


Once optimized, $\mathbf{z}_{\mathrm{test}}^{*}$ is reused to condition the world model for actions that were not observed during latent optimization. Given a different action $\mathbf{a}'$, we predict its outcome as
\begin{equation}
\label{eq:cross_action_prediction}
    \hat{\mathbf{y}}'
    \sim
    p_{\theta^{*}}
    \left(
        \mathbf{y}
        \mid
        \mathbf{x},
        \mathbf{a}',
        \mathbf{z}_{\mathrm{test}}^{*}
    \right).
\end{equation}
  The resulting conditioned world model serves as an environment-adapted test-time simulator, allowing it to predict the consequences of different actions under the inferred environment.

  Unlike amortized history-encoder approaches, \model{} uses direct optimization. This avoids committing to a fixed history mapping and searches for the representation that best explains the current environment. The formulation can naturally incorporate any number of test-time trajectories that differs from training, by optimizing a latent against all available evidence. Moreover, \model{} is not restricted to the range of factor values during training, providing greater flexibility under out-of-distribution configurations. 
  Ablation in \cref{fig:ablation} shows this optimization remains stable across varying test-time trajectory numbers, and is interaction-efficient with a single trajectory sufficient in most settings. \looseness=-1
\section{Experiments}
\label{sec:experiments}

\subsection{Experimental Setup}
\label{sec:experimental_setup}

\paragraph{Tasks}
We evaluate \model{} across 7 simulated settings and 4 real-world
robotic tasks. In simulation, \textbf{Friction}
requires predicting how far an object moves after a push under different
surface friction; \textbf{Gravity} varies gravitational strength and
requires predicting the object motion; \textbf{Mass Collision} varies colliding object mass and predicts the resulting collision motion. \textbf{Light Switch} assigns
different human-defined hidden interaction rules that determine whether a robot action
turns the light on or off; \textbf{Mass Balance} predicts the tilt of a dumbbell under different mass distribution; \textbf{Mass $\times$
	Friction} jointly varies object mass and surface friction at the same time; \textbf{Diverse Visual Backgrounds} evaluates the same friction variations across 30 diverse scene setups, testing whether the learned environment representation remains consistent across visual changes.. In the real world, \textbf{Door Close} varies the spring-lock stiffness,
which changes how the door responds to closing actions; \textbf{Ball
	Friction} requires predicting the motion of a pushed ball of unknown mass; \textbf{Stick Balance} requires predicting the motion of a stick with objects of different masses attached to its ends; and \textbf{Soft Pull}
requires predicting the deformation a soft object under
different rotating actions. 

\paragraph{Implementation Details.}
We adapt ~\cite{bwm2026}, an action-conditioned world model built on Wan2.2-TI2V-5B. The environment latent has a dimension of 32.
Training starts with 300 world-model updates,
followed by 200 latent updates and alternating 200-200 updates.
Stage I/II use LR $10^{-5}$ / $0.03$ respectively. At test time, we
freeze the world model and use a coarse-to-fine LR schedule of
$3.0/1.5/0.5/0.15$, 10 steps each, for 40 steps. \model{} uses only one test-time trajectory on randomly-sampled action for Friction, Gravity, Mass Collision,
Mass $\times$ Friction, Ball Friction, and Soft Pull, and two for Light
Switch, Mass Balance, Door Close, and Stick Balance. \looseness=-1

\begin{table*}[t]
	\centering
	\caption{\small \textbf{Simulation performance across environmental variations.}
		For Friction, Gravity, Mass Collision, Mass $\times$
		Friction, and Diverse Scenes, physical error is measured in pixels; for Light Switch, it is
		lamp-score error; and for Mass Balance, it is bar-tilt error in degrees.
		The average is computed over all seven tasks.}
	\label{tab:sim_main}
	\begingroup
	\scriptsize
	\setlength{\tabcolsep}{2pt}
\renewcommand{\arraystretch}{0.92}
	\begin{tabular*}{\linewidth}{@{\extracolsep{\fill}}llccccc@{}}
		\toprule
		Task & Method
		& \shortstack{Physical Error (Mean) $\downarrow$}
		& \shortstack{Physical Error (Final) $\downarrow$}
		& LPIPS $\downarrow$
		& PSNR $\uparrow$
		& \shortstack{Action Success $\uparrow$} \\
		\midrule
		
		\multirow{5}{*}{Mass Balance}
		& Standard World Model
		& 1.1844 & 4.5571 & 0.0244 & 32.969 & 40.0\% \\
		& LoRA TTA
		& 1.7975 & 7.3105 & 0.0631 & 24.895 & 50.0\% \\
		& DINOv2 Context
		& 1.2636 & 4.9480 & \textbf{0.0239} & \textbf{33.263} & 30.0\% \\
		& TTT-KQV
		& 2.5075 & 10.1759 & 0.0312 & 32.114 & 30.0\% \\
		& \textbf{\model{} (Ours)}
		& \textbf{0.6403} & \textbf{2.0842} & 0.0240 & 32.814 & \textbf{80.0\%} \\
		
		\midrule
		
		\multirow{5}{*}{Friction}
		& Standard World Model
		& 11.0626 & 24.4116 & 0.0667 & 31.113 & 32.0\% \\
		& LoRA TTA
		& 8.1902 & 19.1902 & 0.0582 & \textbf{32.031} & 48.0\% \\
		& DINOv2 Context
		& 11.8142 & 26.1703 & 0.0681 & 30.697 & 48.0\% \\
		& TTT-KQV
		& 10.8137 & 23.8675 & 0.0649 & 30.779 & 32.0\% \\
		& \textbf{\model{} (Ours)}
		& \textbf{3.8540} & \textbf{7.5931} & \textbf{0.0495} & 31.976 & \textbf{72.0\%} \\
		
		\midrule
		
		\multirow{5}{*}{Gravity}
		& Standard World Model
		& 21.4246 & 34.8779 & 0.0345 & 32.509 & 23.5\% \\
		& LoRA TTA
		& 7.9005 & 13.7137 & 0.0231 & 33.328 & 47.1\% \\
		& DINOv2 Context
		& \textbf{5.9300} & \textbf{8.8290} & \textbf{0.0195} & \textbf{34.095} & 76.5\% \\
		& TTT-KQV
		& 18.9572 & 30.9355 & 0.0291 & 33.308 & 35.3\% \\
		& \textbf{\model{} (Ours)}
		& 7.3408 & 11.0653 & 0.0202 & 33.389 & \textbf{82.4\%} \\
		
		\midrule
		
		\multirow{5}{*}{Mass Collision}
		& Standard World Model
		& 21.9329 & 25.7364 & 0.0904 & 29.805 & 38.9\% \\
		& LoRA TTA
		& 15.8704 & 24.5877 & 0.0691 & 30.049 & 33.3\% \\
		& DINOv2 Context
		& 8.5264 & 10.2101 & 0.0540 & 31.039 & 38.9\% \\
		& TTT-KQV
		& 20.9753 & 29.2930 & 0.0842 & 29.385 & 55.6\% \\
		& \textbf{\model{} (Ours)}
		& \textbf{6.4522} & \textbf{8.1321} & \textbf{0.0538} & \textbf{31.150} & \textbf{61.1\%} \\
		
		\midrule
		
		\multirow{5}{*}{Light Switch}
        & Standard World Model
        & 10.3500 & 18.8200 & 0.0187 & 32.619 & 50.0\% \\
        & LoRA TTA
        & 8.2800 & 14.7700 & 0.0175 & 33.123 & 50.0\% \\
        & DINOv2 Context
        & 10.0000 & 18.3700 & 0.0160 & 33.074 & 50.0\% \\
        & TTT-KQV
        & 10.4800 & 18.8700 & 0.0161 & 32.692 & 50.0\% \\
        & \textbf{\model{} (Ours)}
        & \textbf{3.4300} & \textbf{4.7400} & \textbf{0.0150} & \textbf{33.517} & \textbf{87.5\%} \\
		
		\midrule
		
		\multirow{5}{*}{Mass $\times$ Friction}
		& Standard World Model
		& 16.0219 & 29.2454 & 0.0610 & 31.077 & 35.0\% \\
		& LoRA TTA
		& 16.8282 & 26.9302 & 0.0701 & 29.865 & 26.7\% \\
		& DINOv2 Context
		& 11.8116 & 19.0756 & 0.0565 & \textbf{31.367} & 60.8\% \\
		& TTT-KQV
		& 18.9511 & 29.1411 & 0.0726 & 29.739 & 33.3\% \\
		& \textbf{\model{} (Ours)}
		& \textbf{8.1481} & \textbf{13.7222} & \textbf{0.0500} & 30.673 & \textbf{69.7\%} \\
		
		\midrule
		
		\multirow{5}{*}{Diverse Backgrounds}
		& Standard World Model
		& 8.0260 & 16.1831 & 0.1059 & 28.201 & 45.8\% \\
		& LoRA TTA
		& 6.5805 & 16.2328 & 0.1074 & 27.986 & 58.3\% \\
		& DINOv2 Context
		& 7.7367 & 16.0065 & 0.1064 & 28.000 & 58.3\% \\
		& TTT-KQV
		& 7.7783 & 15.5869 & 0.1116 & 27.691 & 58.3\% \\
		& \textbf{\model{} (Ours)}
		& \textbf{3.7145} & \textbf{6.3403} & \textbf{0.0888} & \textbf{28.804} & \textbf{66.7\%} \\
		
		\midrule
		
		\multirow{5}{*}{\textbf{Average}}
        & Standard World Model
        & 12.8575 & 21.9759 & 0.0574 & 31.185 & 37.9\% \\
        & LoRA TTA
        & 9.3496 & 17.5336 & 0.0584 & 30.182 & 44.8\% \\
        & DINOv2 Context
        & 8.1546 & 14.8014 & 0.0492 & 31.648 & 51.8\% \\
        & TTT-KQV
        & 12.9233 & 22.5528 & 0.0585 & 30.815 & 42.1\% \\
        & \textbf{\model{} (Ours)}
        & \textbf{4.7971} & \textbf{7.6682} & \textbf{0.0430} & \textbf{31.760} & \textbf{74.2\%} \\
		
		\bottomrule
	\end{tabular*}
	\endgroup
\end{table*}

\paragraph{Evaluation Metrics.}
(1) \textbf{Physical Error} measures task-relevant state prediction:
object-location error in pixels for Friction, Gravity, Mass Collision, and Diverse Backgrounds; lamp activation error for Light Switch; and bar-tilt
angle error in degrees for Mass Balance. For real-world tasks, we measure
tracked-state error in pixels. We report \textbf{Physical Error (Mean)},
averaged over predicted frames, and \textbf{Physical Error (Final)}, measured
at the final predicted frame.
(2) \textbf{Video Generation Quality} is measured by LPIPS and PSNR between predicted and ground-truth videos. LPIPS measures perceptual discrepancy in a learned feature space, capturing differences in visual structure and appearance, and serves as our primary measure of generation quality. In contrast, PSNR primarily reflects pixel-level reconstruction fidelity and is less informative about semantically meaningful differences or whether object motion is physically correct.
(3) \textbf{Action Success} treats the world model as a simulator: given candidate actions, we predict their outcomes and select the action most likely to achieve the task goal, such as pushing a ball to a target, balancing a stick, or closing a door. We execute the chose action in real world and observe if meets the success criterions in Supp \cref{app:sim}.

\paragraph{Baselines.}
We compare \model{} with four baselines using the same world-model backbone. \textbf{Standard World Model} uses no environment latent or test-time adaptation; \textbf{LoRA TTA} adapts attention LoRA parameters from test-time interactions; \textbf{DINOv2 Context} amortizes a 32-D environment context from interaction trajectories; and \textbf{TTT-KQV} adapts fast weights within the world model. All adaptive baselines receive the same test-time interaction trajectories as \model{}.

Please refer to the Supplementary for more details about architecture (\cref{app:architecture}), task definitions (\cref{app:sim}), baseline implementation (\cref{app:baselines}), and compute efficiency comparison (\cref{app:efficiency}). \looseness=-1

\subsection{Simulation Experiments}
\label{sec:simulation}

\paragraph{Single-Factor Environmental Variation.}
We first evaluate \model{} under individual unobserved environmental factors, spanning
continuous physical properties, mass-dependent interactions, and discrete
human-defined rules. As shown in
\cref{tab:sim_main}, \model{} achieves the highest action success on all five
tasks, increasing the five-task average from $36.9\%$ for the Standard World
Model to $76.6\%$, while reducing average LPIPS from $0.0469$ to $0.0325$.
It also substantially improves task-relevant prediction, reducing mean error
from $11.06$ to $3.85$ pixels on Friction, $21.93$ to $6.45$ pixels on Mass
Collision, and $1.18^\circ$ to $0.64^\circ$ on Mass Balance. These gains arise because \model{} directly optimizes an environment latent from observed interactions, capturing environment-specific variation more flexibly than non-adaptive or amortized baselines.

\begin{figure*}[htb]
	\centering
	\includegraphics[width=0.95\linewidth]{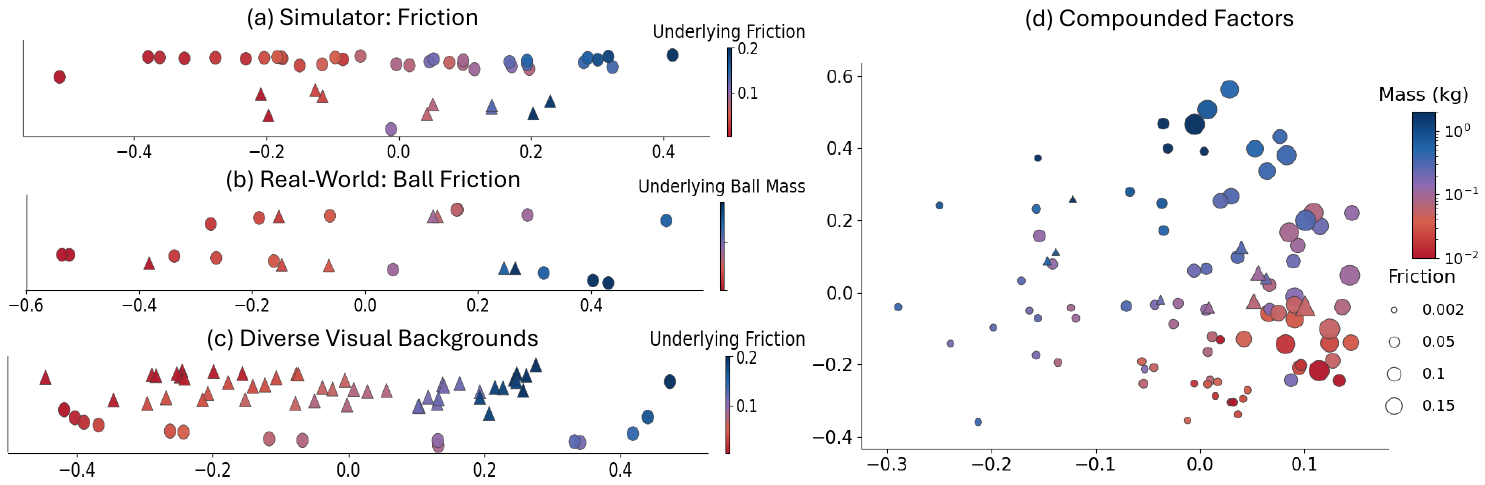}
	\caption{\small \textbf{PCA of learned environment latents.}
		We visualize learned environment latents using PCA onto x-y axes, with
		$\bigcirc$ denoting training environments and $\triangle$ denoting test
		environments. Ground-truth factors (denoted by color or size) are shown only for visualization and are
		never provided to the model.
		\textbf{(a--b)} show that structured latent representations emerge in both
		simulated and real-world single-factor settings.
		\textbf{(c)} shows that a unified latent structure emerges across diverse
		visual backgrounds, indicating that the representation does not become
		entangled with scene appearance.
		\textbf{(d)} shows that a two-dimensional latent structure emerges when mass
		and friction vary jointly, with x-axis primarily corresponding to
		friction and y-axis to mass, despite the model never knowing how many
		underlying factors are present or what they represent.
	}
	\label{fig:pca_latent}
\end{figure*}
\paragraph{Emergent Structure of Environment Latents.}
We examine whether direct optimization gives
rise to structured environment representations without supervision of the
identity or value of underlying factors. In \cref{fig:pca_latent}, we apply principal component
analysis (PCA) to the learned environment latents and reduce them to two
dimensions for visualization. We annotate the projected points using their
corresponding ground-truth factor values to reveal their relationship with
the learned representation; these values are never provided to the model.
\Cref{fig:pca_latent}(a) shows the single-factor friction experiment.
The learned latents form a
one-dimensional structure for a single factor that corresponds closely to the underlying
friction values, as reflected by the point colors: similar values are
located nearby, while friction varies consistently x-axis. PCA visualizations for other tasks are provided in the Supplementary Material \cref{app:pca}.

\paragraph{Compounded Environmental Factors.}
We next evaluate \model{} under the more challenging setting of compounded
environmental factors, where object mass and surface friction vary
simultaneously and jointly
reflect in object motion. Importantly, \model{} is never informed that
there are two underlying factors, nor given their identities or values.
\cref{tab:sim_main} shows \model{} achieves the lowest
physical errors, while improving action success from
$35.0\%$ to $69.7\%$. More strikingly, \cref{fig:pca_latent}(d) shows that
the learned latents naturally form a two-dimensional structure corresponding
to the two underlying factors, showing that \model{} discovers the structure
of compounded environmental factors through prediction alone, without
knowing in advance how many factors are present or what they represent. \looseness=-1

\paragraph{Robustness across Diverse Visual Scenes.}
We test whether the learned environment representation captures
underlying dynamics without becoming entangled with visual appearance.
Specifically, we ask whether a unified latent structure emerges across
visually diverse scenes despite substantial differences in appearance. We construct a multi-background setting in which the same underlying
friction factor varies across diverse scene layouts and visual backgrounds.
\cref{tab:sim_main} shows that \model{} outperforms all baselines across every metric. \cref{fig:pca_latent}(c) further shows that latents from
different backgrounds organize consistently rather than separating according to scene
appearance. This indicates that environment-latent can isolate
environment-specific information that is shared across visually different
observations, enabling a unified latent structure to emerge across diverse
scenes. \looseness=-1

		
		
		
		

\begin{table*}[t]
	\centering
	\caption{\small \textbf{Real-world robotic task performance.}
		\model{} outperforms baselines across real-world tasks.}
	\label{tab:real_world}
	\begingroup
	\scriptsize
	\setlength{\tabcolsep}{2pt}
\renewcommand{\arraystretch}{0.92}
	
	\begin{tabular*}{\linewidth}{@{\extracolsep{\fill}}llccccc@{}}
		\toprule
		Task & Method
		& \shortstack{Physical Error (Mean) $\downarrow$}
		& \shortstack{Physical Error (Final) $\downarrow$}
		& LPIPS $\downarrow$
		& PSNR $\uparrow$
		& \shortstack{Action Success $\uparrow$} \\
		\midrule
		
		\multirow{5}{*}{Door Close}
		& Standard World Model
		& 21.82 & 41.22 & 0.05700 & 29.96 & 22.22\% \\
		& LoRA TTA
		& 17.35 & 32.14 & 0.04946 & 30.24 & 22.22\% \\
		& DINOv2 Context
		& 21.12 & 38.90 & 0.05490 & 30.28 & 22.22\% \\
		& TTT-KQV
		& 19.87 & 35.93 & 0.05164 & 30.95 & 22.22\% \\
		& \textbf{\model{} (Ours)}
		& \textbf{7.95} & \textbf{13.47} & \textbf{0.03208}
		& \textbf{32.15} & \textbf{55.56\%} \\
		
		\midrule
		
		\multirow{5}{*}{Ball Friction}
		& Standard World Model
		& 28.36 & 64.42 & 0.03153 & 36.04 & 0.00\% \\
		& LoRA TTA
		& 37.48 & 72.09 & 0.03805 & 35.44 & 50.00\% \\
		& DINOv2 Context
		& 24.90 & 52.54 & 0.02934 & \textbf{36.32} & 12.50\% \\
		& TTT-KQV
		& 29.80 & 67.37 & 0.03224 & 36.29 & 25.00\% \\
		& \textbf{\model{} (Ours)}
		& \textbf{14.47} & \textbf{24.57} & \textbf{0.02401}
		& 36.23 & \textbf{62.50\%} \\
		
		\midrule
		
		\multirow{5}{*}{Stick Balance}
		& Standard World Model
		& 5.29 & 11.40 & 0.04716 & 31.87 & 20.00\% \\
		& LoRA TTA
		& 3.63 & 7.24 & 0.04111 & 32.40 & 40.00\% \\
		& DINOv2 Context
		& 2.80 & 7.54 & \textbf{0.03640} & \textbf{33.58} & 20.00\% \\
		& TTT-KQV
		& 4.49 & 10.40 & 0.03948 & 33.38 & 25.00\% \\
		& \textbf{\model{} (Ours)}
		& \textbf{2.27} & \textbf{5.01} & 0.04175
		& 32.05 & \textbf{83.33\%} \\
		
		\midrule
		
		\multirow{5}{*}{Soft Pull}
		& Standard World Model
		& 21.71 & 35.90 & 0.07207 & 33.32 & 63.64\% \\
		& LoRA TTA
		& 18.63 & 26.20 & 0.06862 & 33.63 & 63.64\% \\
		& DINOv2 Context
		& 21.19 & 29.17 & 0.07087 & 33.47 & 72.73\% \\
		& TTT-KQV
		& \textbf{18.23} & 29.69 & \textbf{0.06439} & \textbf{34.03} & 72.73\% \\
		& \textbf{\model{} (Ours)}
		& 18.71 & \textbf{23.59} & 0.07580
		& 32.52 & \textbf{81.82\%} \\
		
		\midrule
		
		\multirow{5}{*}{\textbf{Average}}
		& Standard World Model
		& 19.30 & 38.24 & 0.05194 & 32.80 & 26.47\% \\
		& LoRA TTA
		& 19.27 & 34.42 & 0.04931 & 32.93 & 43.97\% \\
		& DINOv2 Context
		& 17.50 & 32.04 & 0.04788 & 33.41 & 31.86\% \\
		& TTT-KQV
		& 18.10 & 35.85 & 0.04694 & \textbf{33.66} & 36.24\% \\
		& \textbf{\model{} (Ours)}
		& \textbf{10.85} & \textbf{16.66} & \textbf{0.04341}
		& 33.24 & \textbf{70.80\%} \\
		
		\bottomrule
	\end{tabular*}
	\endgroup
\end{table*}


\subsection{Real-World Robotic Experiments}
\label{sec:real_world}

We further evaluate \model{} on four real-world tasks. As shown in
\cref{tab:real_world}, compared with the Standard World Model, \model{}
reduces average mean/final physical error from $19.30/38.24$ to
$10.85/16.66$ pixels and LPIPS from $0.05194$ to $0.04341$, while increasing
action success from $26.47\%$ to $70.80\%$. On Stick Balance, for example,
action success increases from $20.00\%$ to $83.33\%$. 
\cref{fig:pca_latent}(b) reveals a clear one-dimensional latent structure
without ground-truth friction supervision or post-hoc annotation. \model{} is particularly important in the real world, where underlying environmental factors are typically unknown and unmeasured and interactions contain additional uncontrolled variation. Notably, \model{} achieves even larger improvements over the Standard World Model than in simulation, indicating greater benefits as environmental variation becomes less controlled.
\begin{figure*}[htbp]
	\centering
	\includegraphics[width=\linewidth]{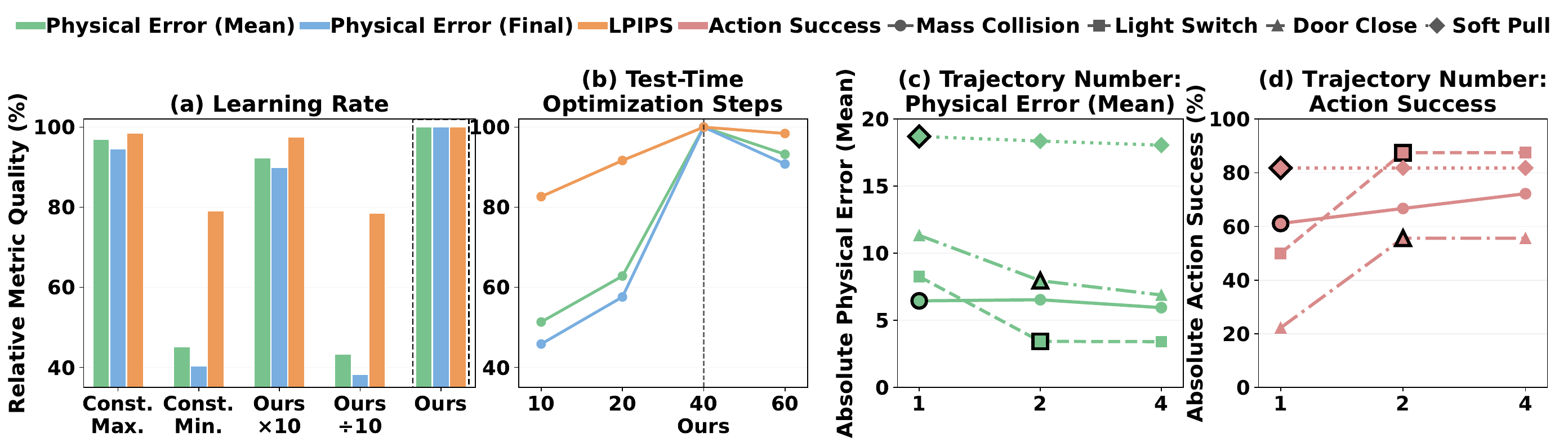}
	\caption{\small\textbf{Ablation studies.}
		\textbf{(a)} Test-Time Optimization Learning-rate strategy.
		\textit{Constant Maximum/Minimum} remove coarse-to-fine scheduling, while
\textit{Ours $\times10/\div10$} scale our schedule uniformly.
		\textbf{(b)} Ablating the test-time optimization budget shows that performance peaks at 40 steps.
		\textbf{(c,d)} Effect of the number of test-time interaction trajectories. Bold markers denote
		our chosen trajectory count, after which performance largely saturates. \looseness=-1
	}
	\label{fig:ablation}
\end{figure*}
\subsection{Ablative \& Qualitative Studies}
\label{sec:analysis}

We ablate the key design choices of \model{} in \cref{fig:ablation} and
\cref{tab:ablation}, including latent optimization, interaction evidence,
latent parameterization, and alternating training.
In \cref{fig:ablation}(a), we vary the learning-rate strategy.
 Our schedule
performs best, validating both its coarse-to-fine form and learning-rate
scale. \Cref{fig:ablation}(b) varies test-time optimization from 10--60
steps; performance improves up to 40 steps and then saturates, motivating
our default. \Cref{fig:ablation}(c,d) ablate the number of test-time
interaction trajectories.
Performance largely saturates beyond the selected number of trajectories,
indicating that additional interaction provides limited benefit. Recall that
\model{} uses only one test-time trajectory for 7 tasks, and two for 4 tasks, demonstrating that
\model{} can infer useful environment information with as few as one or two
interactions.
 Additional ablations in \cref{tab:ablation} study latent parameterization
and training. All the ablated models underperform our model,
supporting our alternating schedule and latent mechanism. We provide the latent-dimensionality ablation in Supplementary \cref{tab:supp-latent-dim}.

In \cref{fig:qualitative}, we show qualitative results across simulation and
real-world tasks. 
\model{}
closely matches ground-truth outcomes, indicating
that the inferred latent successfully conditions prediction on the current
environment. By contrast, the baselines fail to capture these
environment-specific differences, producing largely random motions and
incorrect final configurations across environments. Please refer to the Supplementary Material for more qualitative examples (\cref{app:qualitative}).

\begin{table*}[htbp]
	\centering
	\caption{\small\textbf{Additional Ablation study.}
		\textit{Joint Model--Latent Training} optimizes the world model and latent
		together without separate stages; \textit{Direct Token} directly
		optimizes the full-dimensional conditioning token; \textit{32-D Random
			$[-0.05,0.05]$} changes the latent initialization range; and
		\textit{Alternating Latent/Model} change
		the number of updates before switching between the two optimization stages,
		compared with our default 200/200.}
	\label{tab:ablation}
	\begingroup
	\scriptsize
	\setlength{\tabcolsep}{2pt}
	\renewcommand{\arraystretch}{0.92}
	\begin{tabular*}{\linewidth}{@{\extracolsep{\fill}}lccccc@{}}
		\toprule
		Method
		& \shortstack{Physical Error (Mean) $\downarrow$}
		& \shortstack{Physical Error (Final) $\downarrow$}
		& LPIPS $\downarrow$
		& PSNR $\uparrow$
		& \shortstack{Action Success $\uparrow$} \\
		\midrule
		
		Joint Model-Latent Training
		& 7.03 & 15.66 & 0.0657 & 31.203 & 68\% \\
		
		Direct Token (3072-D)
		& 94.50 & 80.49 & 0.7626 & 15.218 & 24\% \\
		
		32-D Random $[-0.05,0.05]$
		& 4.81 & 9.73 & 0.0495 & 32.172 & 60\% \\
		
		Alternating Latent/Model (100/100)
		& 5.41 & 11.50 & \textbf{0.0484} & 32.208 & 56\% \\
		Alternating Latent/Model (400/400)
		& 4.31 & 8.80 & 0.0507 & \textbf{32.430} & 68\% \\
		
		
		\textbf{\model{}}
		& \textbf{3.85} & \textbf{7.59} & 0.0495 & 31.976 & \textbf{72\%} \\
		
		\bottomrule
	\end{tabular*}
	\endgroup
\end{table*}

\vspace{-1em}
\begin{figure*}[htbp]
	\centering
	\includegraphics[width=\linewidth]{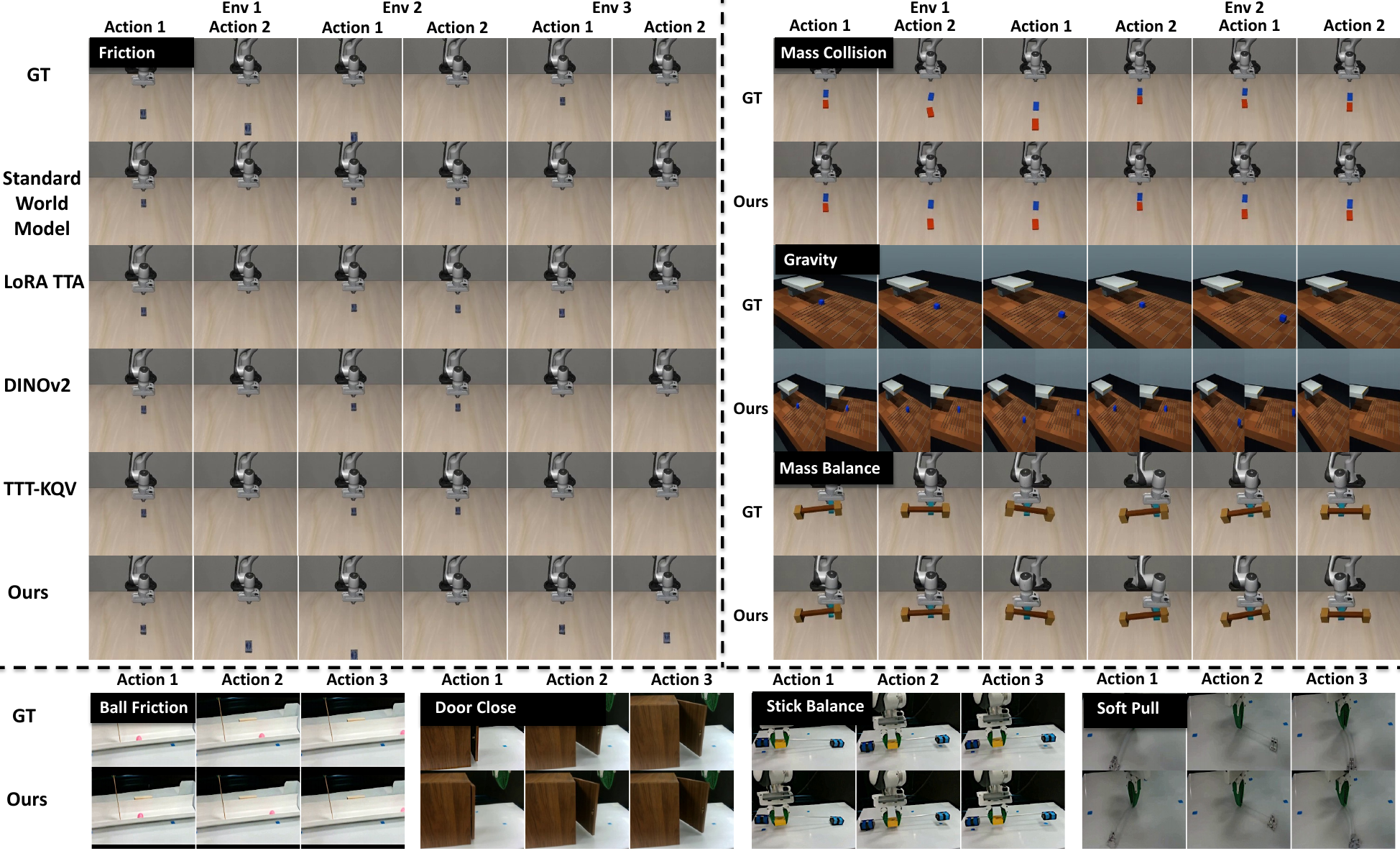}
    \vspace{-2em}
	\caption{\small\textbf{Qualitative results in simulation and the real world.}
Within each task, we apply the same actions across different environments,
which produce different ground-truth outcomes due to underlying environmental
variation. \model{} closely matches these environment-dependent outcomes.}
	\label{fig:qualitative}
\end{figure*}

\section{Conclusion}
We introduce \model{} which augments robotic world models with enviornment latent conditioning.  
\model{} alternates between world-model optimization that learns action-conditioned dyanmics, and environment-latent optimization that captures hidden environment factors. For a new environment, the world model is frozen and only the latent is optimized.
Across 7 simulation and 4 real-world tasks, \model{} outperforms baselines a lot in both video prediction and action selection. 

\section*{Acknowledgements}
This paper is in part funded by ONR MURI N00014-24-1-2748, ONR MURI N00014-22-1-2740, Singapore MDDI AIVP-2025-008, IITP funded by the Korean Government (MSIT) (No. RS-2024-00457882, National AI Research Lab Project), 2077AI, and Grant AW1134392 (Reasoning in Motion) from the TRI University 3.0 Program. 



\bibliography{iclr2027_conference}

\addcontentsline{tzc}{section}{References}
\begin{thebibliography}{5}
\providecommand{\natexlab}[1]{#1}

\bibitem[{BWM Team}(2026)]{supp:bwm2026}
{BWM Team}.
\newblock {BWM}: A low-cost high-fidelity world simulator for robot learning.
\newblock \emph{arXiv preprint arXiv:2607.29302}, 2026.

\bibitem[Dalal et~al.(2025)Dalal, Koceja, Xu, Zhao, Han, Cheung, Kautz, Choi,
  Sun, and Wang]{supp:dalal2025oneminute}
Karan Dalal, Daniel Koceja, Jiarui Xu, Yue Zhao, Shihao Han, Ka~Chun Cheung,
  Jan Kautz, Yejin Choi, Yu~Sun, and Xiaolong Wang.
\newblock One-minute video generation with test-time training.
\newblock In \emph{Proceedings of the IEEE/CVF Conference on Computer Vision
  and Pattern Recognition}, pp.\ 17702--17711, 2025.

\bibitem[Hu et~al.(2022)Hu, Shen, Wallis, Allen-Zhu, Li, Wang, Wang, and
  Chen]{supp:lora}
Edward~J. Hu, Yelong Shen, Phillip Wallis, Zeyuan Allen-Zhu, Yuanzhi Li, Shean
  Wang, Lu~Wang, and Weizhu Chen.
\newblock {LoRA}: Low-rank adaptation of large language models.
\newblock In \emph{International Conference on Learning Representations}, 2022.

\bibitem[Liu et~al.(2023)Liu, Zhu, Gao, Feng, Liu, Zhu, and
  Stone]{supp:liu2023libero}
Bo~Liu, Yifeng Zhu, Chongkai Gao, Yihao Feng, Qiang Liu, Yuke Zhu, and Peter
  Stone.
\newblock {LIBERO}: Benchmarking knowledge transfer for lifelong robot learning.
\newblock In \emph{Advances in Neural Information Processing Systems, Datasets
  and Benchmarks Track}, 2023.

\bibitem[Oquab et~al.(2024)Oquab, Darcet, Moutakanni, Vo, Szafraniec, Khalidov,
  Fernandez, Haziza, Massa, El-Nouby, Assran, Ballas, Galuba, Howes, Huang, Li,
  Misra, Rabbat, Sharma, Synnaeve, Xu, J{\'e}gou, Mairal, Labatut, Joulin, and
  Bojanowski]{supp:dinov2}
Maxime Oquab, Timoth{\'e}e Darcet, Th{\'e}o Moutakanni, Huy~V. Vo, Marc
  Szafraniec, Vasil Khalidov, Pierre Fernandez, Daniel Haziza, Francisco Massa,
  Alaaeldin El-Nouby, Mido Assran, Nicolas Ballas, Wojciech Galuba, Russell
  Howes, Po-Yao Huang, Shang-Wen Li, Ishan Misra, Michael Rabbat, Vasu Sharma,
  Gabriel Synnaeve, Hu~Xu, Herv{\'e} J{\'e}gou, Julien Mairal, Patrick Labatut,
  Armand Joulin, and Piotr Bojanowski.
\newblock {DINOv2}: Learning robust visual features without supervision.
\newblock \emph{Transactions on Machine Learning Research}, 2024.

\end{thebibliography}
\bibliographystyle{iclr2027_conference}

\clearpage
\appendix

\makeatletter
\newcommand{\tztoc}{\@starttoc{tzc}}
\makeatother
\newcommand{\tzsection}[1]{%
  \section{#1}%
  \addcontentsline{tzc}{section}{\protect\numberline{\thesection}#1}}
\newcommand{\tzsubsection}[1]{%
  \subsection{#1}%
  \addcontentsline{tzc}{subsection}{\protect\numberline{\thesubsection}#1}}

\begin{center}
{\Large\bfseries Supplementary Material for}\\[0.3em]
{\large\bfseries T2-Z: Test-Time Optimization of Environment Latents\\
for Robotic World Models}
\end{center}
\section*{Contents}
{\hypersetup{linkcolor=black}\tztoc}
\clearpage

\tzsection{Model Architecture and Optimization}
\label{app:architecture}

\tzsubsection{Architecture and latent conditioning}
\paragraph{Inputs and inherited backbone.}
For an episode in environment $e$, the model receives one initial RGB frame
$I_0$ and an aligned sequence $a=(a_t)_t$ of 14-D robot actions, and predicts
the subsequent video frames. We initialize the video diffusion transformer
(DiT) and action encoder from BWM~\citep{bwm2026}, retaining its
Wan2.2\citep{wan2025}-based action-conditioning pathway. The frozen video VAE encodes the
initial frame into the video-latent stream and decodes predicted latents
back to RGB. Text conditioning is disabled; the initial frame is not an
extra cross-attention token.
In short, $I_0$ enters the video-latent stream, $a$ follows BWM's two
action-conditioning paths, and the environment latent augments both paths.

\paragraph{Environment-indexed latent conditioning.}
We associate each environment $e$ with a freely optimized
vector $z_e\in\mathbb{R}^{32}$. Episodes assigned to an environment share its latent; their
initial frames and actions vary, so predicting their different outcomes
constrains the environment representation. There is no
observation-to-latent encoder and no supervision
from the numerical physical parameter. Let $D=3072$ be the DiT width.
Two separate projectors turn $z_e$ into model-width vectors:
\begin{equation}
 \begin{aligned}
 q_b(z_e)&=W_{b,2}\,\mathrm{GELU}\!\left(W_{b,1}\,\mathrm{LN}(z_e)\right),
 \quad b\in\{\mathrm{token},\mathrm{mod}\},\\
 W_{b,1}&:32\to128,\qquad W_{b,2}:128\to D.
 \end{aligned}
 \label{eq:supp-injection}
\end{equation}
The two linear layers in each projector have no bias; the projectors do
not share weights. The 128-D width is internal to these MLPs: both outputs
are $D$-dimensional.

\paragraph{Cross-attention branch.}
BWM's action encoder maps each action to a $D$-dimensional conditioning
token. Let $A_{\mathrm{tok}}(a)$ denote this token sequence and
$u_e=q_{\mathrm{token}}(z_e)$ the single environment token. With text
disabled, each DiT block receives the conditioning sequence
$C_e=[u_e;A_{\mathrm{tok}}(a)]$. For video patch tokens
$X_\ell\in\mathbb{R}^{N\times D}$ entering block $\ell$, the attention
operation can be written schematically as
\begin{equation}
 \operatorname{CA}(X_\ell,C_e)=
 \operatorname{softmax}\!\left(
 \frac{(X_\ell W_Q)(C_eW_K)^\top}{\sqrt{d_h}}\right)
 (C_eW_V)W_O.
 \label{eq:supp-cross-attn}
\end{equation}
Here $N$ is the number of video patches, $d_h$ is one attention head's
width, and $W_Q,W_K,W_V,W_O$ are the block's learned attention matrices.
The actual operation is multi-head; the equation shows where the new
token enters. Every video patch can attend to the environment token as
well as the action tokens through keys and values.

\paragraph{Modulation branch.}
BWM also groups four aligned 14-D actions at a time and maps each group
to a $D$-dimensional action-modulation vector $a_g^{\mathrm{mod}}$.
Index $g$ identifies the corresponding latent-video temporal group, and
$s$ identifies a spatial patch within it. Let $\tau$ be the current
flow-matching timestep and $b_{g,s}(\tau)\in\mathbb{R}^{D}$ be Wan's
\emph{base timestep embedding} at that video patch, before action or
environment conditioning. The initial-frame latent uses the unnoised
timestep, while predicted-frame latents use $\tau$. The environment-modulation vector is
$v_e=q_{\mathrm{mod}}(z_e)\in\mathbb{R}^{D}$. The implementation adds
these three vectors and applies the existing time projection:
\begin{equation}
 \tilde b_{g,s}(\tau)=b_{g,s}(\tau)+a_g^{\mathrm{mod}}+v_e,
 \qquad M_{g,s}=\operatorname{TimeProj}(\tilde b_{g,s}(\tau))
 \in\mathbb{R}^{6\times D}.
 \label{eq:supp-modulation}
\end{equation}
The action and environment vectors are broadcast to every spatial patch
in group $g$. The six $D$-dimensional outputs of $M_{g,s}$ supply the
DiT blocks' adaptive-normalization shifts, scales, and residual gates;
the modified timestep embedding also conditions the output head. Thus
$z_e$ influences both attention and per-patch modulation, while leaving
BWM's original action pathways in place.

\begin{table}[htbp]
\centering\small
\caption{Model components and optimization roles. Test-time adaptation
updates only the environment latent.}
\label{tab:supp-components}
\begin{tabular}{@{}p{0.25\linewidth}p{0.43\linewidth}p{0.25\linewidth}@{}}
\toprule
Component & Initialization or structure & Training update \\
\midrule
Video DiT & BWM-initialized Wan2.2 backbone & Model blocks \\
Action encoder & BWM-initialized 14-D action pathway & Model blocks \\
Video VAE & Wan2.2 video encoder/decoder & Frozen \\
Environment table & Learnable 32-D latents indexed by environment & Latent blocks \\
Latent projectors & Two $32\to128\to3072$ MLPs & Model blocks \\
\bottomrule
\end{tabular}
\end{table}

\tzsubsection{Training and test-time adaptation}
\paragraph{Training schedule.}
Training uses the inherited flow-matching video-prediction loss, excluding
the initial conditioning frame. A latent block freezes the shared model and
updates the environment table; a model block freezes the table and updates the
DiT, action encoder, and two projectors. The basic cycle alternates a
200-update latent block and a 200-update model block. An initial 300-update model block precedes
the curriculum. The shared-model optimizer is AdamW with learning rate
$10^{-5}$, weight decay $0.01$, 100 warmup updates, gradient clipping $0.5$,
and bfloat16 training. The environment-latent optimizer uses learning
rate $0.03$ and zero weight decay. Latents are independently initialized from
$\mathcal U(0,1)$ and clipped to $[0,1]$ after training updates.

\paragraph{Test-time adaptation.}
For a held-out environment, a 32-D latent is optimized from its
test-time interaction trajectories; all shared weights stay frozen. Query outcomes never
enter this update. Both simulation and real-world main evaluations use
40 latent updates with learning rates $3.0,1.5,0.5,0.15$, each for ten
updates. The resulting latent is reused across query actions from that
environment. Exact task-specific execution configurations are released with
the code.

\tzsection{Experimental Setup}
\tzsubsection{Training data and resource budgets}
\label{app:training}

Tables~\ref{tab:supp-sim-train} and~\ref{tab:supp-real-train} report counts of
physical environments and recorded episodes, rather than latent entries.
Each video chunk contains
one initial conditioning frame and the remaining frames are prediction
targets. The batch notation is \emph{training environments $\times$ action
chunks per environment} per optimizer update.

\begin{table}[htbp]
\centering\small
\setlength{\tabcolsep}{4pt}
\caption{Simulation training scale by physical environment. Diverse Backgrounds uses 30 scene
backgrounds. Budget is H200 GPUs $\times$ hours per method.}
\label{tab:supp-sim-train}
\begin{tabular}{@{}lrrrrr@{}}
\toprule
Task & Physical envs. $\times$ episodes/env. & Frames & Stride & Batch & Budget \\
\midrule
Friction & $35\times10$ & 41 & 1 & $8\times4$ & $2\times24$ \\
Gravity & $20\times10$ & 61 & 1 & $8\times4$ & $2\times24$ \\
Mass Collision & $20\times9$ & 61 & 1 & $8\times8$ & $2\times24$ \\
Light Switch & $12\times10$ & 33 & 1 & $8\times8$ & $2\times24$ \\
Mass Balance & $20\times15$ & 41 & 2 & $8\times8$ & $2\times24$ \\
Mass $\times$ Friction & $84\times9$ & 61 & 1 & $16\times6$ & $4\times48$ \\
Diverse Backgrounds & $15\times50$ & 41 & 1 & $12\times6$ & $4\times24$ \\
\bottomrule
\end{tabular}
\end{table}

Light Switch episodes contain continuous button-press trajectories. Diverse Backgrounds varies friction across physical environments while the
different scene backgrounds provide repeated interactions.

\begin{table}[htbp]
\centering\footnotesize
\setlength{\tabcolsep}{3pt}
\caption{Real-world training scale by physical environment and native/model-input RGB resolution.
Budget is H200 GPUs $\times$ hours per method.}
\label{tab:supp-real-train}
\begin{tabular}{@{}lrrrrrrr@{}}
\toprule
Task & Physical envs. / episodes & Frames & Stride & Native RGB & Input RGB & Batch & Budget \\
\midrule
Ball Friction & 8 / 240 & 61 & 1 & $560\times248$ & $320\times160$ & $6\times8$ & $2\times24$ \\
Door Close & 10 / 438 & 49 & 1 & $640\times480$ & $256\times192$ & $6\times8$ & $2\times24$ \\
Stick Balance & 9 / 421 & 33 & 3 & $640\times480$ & $256\times192$ & $6\times8$ & $2\times24$ \\
Soft Pull & 9 / 146 & 33 & 3 & $512\times256$ & $320\times160$ & $10\times6$ & $2\times24$ \\
\bottomrule
\end{tabular}
\end{table}

\clearpage
\tzsubsection{Simulator and real-world setup}
\label{app:setup}

\paragraph{Simulator.}
We construct the custom tasks in a LIBERO-based simulator~\citep{liu2023libero}
with a Franka Panda arm. Simulated RGB videos are recorded at 20 fps and
$224\times224$ resolution. The action representation has 14 channels.

\paragraph{Real world.}
The physical tasks use a Franka robot arm and a fixed Intel RealSense D435i
camera. We use only its main RGB view, captured at $640\times480$ and
20 fps; the task-specific saved video crops and model-input sizes appear in
Table~\ref{tab:supp-real-train}. Model inputs are letterboxed to the listed
sizes. All tasks use 14-channel action vectors;
the target-action representation is task dependent.

\tzsection{Evaluation Protocols}
\label{app:sim}

\tzsubsection{Shared protocol and metrics}
\paragraph{Shared comparison.}
For action-selection tasks, we sample candidate actions and use each model's
predicted videos to make an action decision. The selected action is then
executed in the environment, and success is determined from its actual
outcome using the task-specific criteria below. Each evaluated candidate-action
set contains at least one action that satisfies the target condition.
All methods receive the same query inputs and candidate actions, and all
methods that adapt at test time receive the same test-time interaction
trajectories. The Standard World Model does not adapt at test time, and no
method sees a query outcome before making its prediction.

For tracked object positions $p_t$ and predictions $\hat p_t$ over $T$
evaluated future frames, average displacement error (ADE) and final
displacement error (FDE) are
\begin{equation}
\mathrm{ADE}=\frac{1}{T}\sum_{t=1}^{T}\|\hat p_t-p_t\|_2,
\qquad
\mathrm{FDE}=\|\hat p_T-p_T\|_2.
\end{equation}
Both are reported in image pixels. Tasks with scalar physical states use
the corresponding mean/final absolute errors stated below. PSNR and SSIM
measure full-frame appearance. LPIPS uses the official pretrained AlexNet
perceptual metric on RGB frames scaled to $[-1,1]$, without an object
mask; per-frame values are averaged over the evaluated future frames and
then over query videos.

A query video denotes one action-conditioned video chunk evaluated at test
time. The counts below sum query videos across the evaluated environments; $K$
is the number of test-time interaction trajectories per test environment.
\begin{table}[htbp]
\centering\small
\setlength{\tabcolsep}{5pt}
\caption{Simulation evaluation counts: test-time interaction trajectories and query videos.}
\label{tab:supp-sim-eval}
\begin{tabular}{@{}lrr@{}}
\toprule
Task & $K$ & Query videos \\
\midrule
Friction & 1 & 90 \\
Gravity & 1 & 90 \\
Mass Collision & 1 & 80 \\
Light Switch & 2 & 60 \\
Mass Balance & 2 & 130 \\
Mass $\times$ Friction & 1 & 80 \\
Diverse Backgrounds & 1 & 90 \\
\bottomrule
\end{tabular}
\end{table}

\tzsubsection{Simulated tasks}

\begin{table}[htbp]
\centering\small
\caption{Simulation object-centric quantities extracted from the predicted
and ground-truth RGB videos. ADE/FDE are in image pixels.}
\label{tab:supp-sim-object}
\begin{tabular}{@{}lp{0.43\linewidth}p{0.25\linewidth}@{}}
\toprule
Task & Tracked quantity & Object-centric error \\
\midrule
Friction & Pushed-block centroid & ADE/FDE \\
Gravity & Blue projectile centroid & ADE/FDE \\
Mass Collision & Red struck-object centroid & ADE/FDE \\
Light Switch & Yellow-pixel fraction in lamp region & Mean/final absolute error (pp) \\
Mass Balance & Balance-bar tilt angle & Mean/final angle error ($^\circ$) \\
Mass $\times$ Friction & Red struck-object centroid & ADE/FDE \\
Diverse Backgrounds & Pushed-block centroid & ADE/FDE \\
\bottomrule
\end{tabular}
\end{table}

\begin{figure}[!ht]
\centering
\begin{minipage}[t]{0.31\linewidth}\centering
\includegraphics[width=\linewidth]{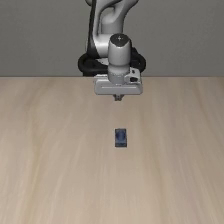}\\
{\footnotesize (a) Friction}
\end{minipage}\hfill
\begin{minipage}[t]{0.31\linewidth}\centering
\includegraphics[width=\linewidth]{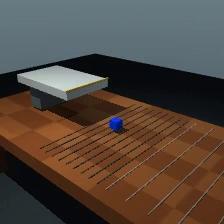}\\
{\footnotesize (b) Gravity}
\end{minipage}\hfill
\begin{minipage}[t]{0.31\linewidth}\centering
\includegraphics[width=\linewidth]{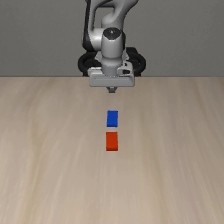}\\
{\footnotesize (c) Mass Collision}
\end{minipage}
\vspace{0.5em}

\begin{minipage}[t]{0.31\linewidth}\centering
\includegraphics[width=\linewidth]{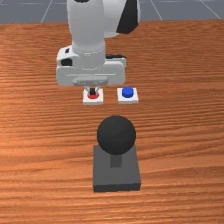}\\
{\footnotesize (d) Light Switch}
\end{minipage}\hfill
\begin{minipage}[t]{0.31\linewidth}\centering
\includegraphics[width=\linewidth]{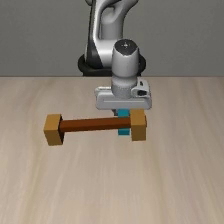}\\
{\footnotesize (e) Mass Balance}
\end{minipage}\hfill
\begin{minipage}[t]{0.31\linewidth}\centering
\includegraphics[width=\linewidth]{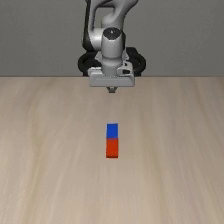}\\
{\footnotesize (f) Mass $\times$ Friction}
\end{minipage}
\caption{Representative ground-truth frames from the simulated tasks.
Diverse Backgrounds uses the Friction push task with varied scene backgrounds.}
\label{fig:supp-sim-tasks}
\end{figure}

\paragraph{Friction.}
We change the table's surface-friction coefficient while the robot pushes
the same block with different action strengths. We track the pushed block's
image centroid. Action Success
means that its observed displacement reaches the requested short, medium,
or long target region.

\paragraph{Gravity.}
We vary the simulator's gravitational acceleration, changing the path of a
blue block after it is thrown. Different actions correspond to different
launch speeds of the block. We track the projectile's centroid. Success means that the
observed projectile endpoint reaches the requested near, middle, or far
target region.

\paragraph{Mass Collision.}
We vary the mass of the red target object while a driver object collides
with it under different actions.
We track the \emph{struck} red object's centroid, not the driver's. Success
means that the struck object reaches the requested displacement region.

\paragraph{Light Switch.}
We vary the hidden causal rule specifying whether the red button, the blue
button, both, or neither can change the lamp state. The object-centric
signal is \emph{color}, not average brightness: a fixed RGB rule detects
yellow pixels in the lamp region $\mathcal R$. Each frame's score is
\begin{equation}
 y_t=\frac{1}{|\mathcal R|}\sum_{p\in\mathcal R}
 \mathbf{1}[\mathrm{RGB}_t(p)\text{ passes the yellow-color rule}].
 \label{eq:supp-light-score}
\end{equation}
We report the absolute difference between predicted and observed scores
in percentage points (pp). For $T$ evaluated future frames,
\begin{equation}
 E_{\mathrm{mean}}^{\mathrm{light}}
 =\frac{100}{T}\sum_{t=1}^{T}|\hat y_t-y_t|,
 \qquad
 E_{\mathrm{final}}^{\mathrm{light}}
 =100|\hat y_T-y_T|.
\end{equation}
A score of at least
$0.35$ denotes ``on'' for the terminal-state goal; success means that the
observed terminal lamp state matches the requested on/off target.

\paragraph{Mass Balance.}
We vary the left/right mass ratio of the balance apparatus; the robot's
support position is the action used to probe that fixed ratio. We track the
bar angle in degrees; its ADE/FDE
analogues are mean/final angular errors. Success means that the observed
terminal bar angle lies in the horizontal balance interval.

\paragraph{Mass $\times$ Friction.}
We combine different red-object masses with different table-friction
coefficients, so both physical factors can change across environments.
We track the red struck
object's centroid. For each target line, the desired action is the least
forceful sampled action whose observed endpoint crosses that line; the
selector estimates this minimum from the predicted videos.

\paragraph{Diverse Backgrounds (multi-background).}
We vary the table-friction coefficient as in Friction while rendering the
same push task under different scene backgrounds; background is a visual
change, not a new action. We again track the pushed block's centroid; success means
reaching the requested displacement region. This tests whether the shared
environment representation follows dynamics rather than appearance.

\tzsubsection{Real-world tests}
\label{app:real}

\begin{table}[htbp]
\centering\small
\setlength{\tabcolsep}{5pt}
\caption{Real-world evaluation counts: test-time interaction trajectories and query videos.}
\label{tab:supp-real-eval}
\begin{tabular}{@{}lrr@{}}
\toprule
Task & $K$ & Query videos \\
\midrule
Ball Friction & 1 & 80 \\
Door Close & 2 & 100 \\
Stick Balance & 2 & 45 \\
Soft Pull & 1 & 24 \\
\bottomrule
\end{tabular}
\end{table}

\begin{table}[htbp]
\centering\small
\caption{Real-world object-centric measurements from the main camera.
ADE/FDE are reported in native-image pixels.}
\label{tab:supp-real-object}
\begin{tabular}{@{}lp{0.48\linewidth}p{0.22\linewidth}@{}}
\toprule
Task & Tracked quantity & Object-centric error \\
\midrule
Ball Friction & Visible ball center & ADE/FDE \\
Door Close & Outer door-edge position & ADE/FDE \\
Stick Balance & Two end markers, averaged & ADE/FDE \\
Soft Pull & Dragged object's visible center & ADE/FDE \\
\bottomrule
\end{tabular}
\end{table}

\begin{figure}[htbp]
\centering
\begin{minipage}[t]{0.48\linewidth}
\centering
\includegraphics[angle=90,trim=650 150 900 150,clip,width=\linewidth,height=0.16\textheight,keepaspectratio]{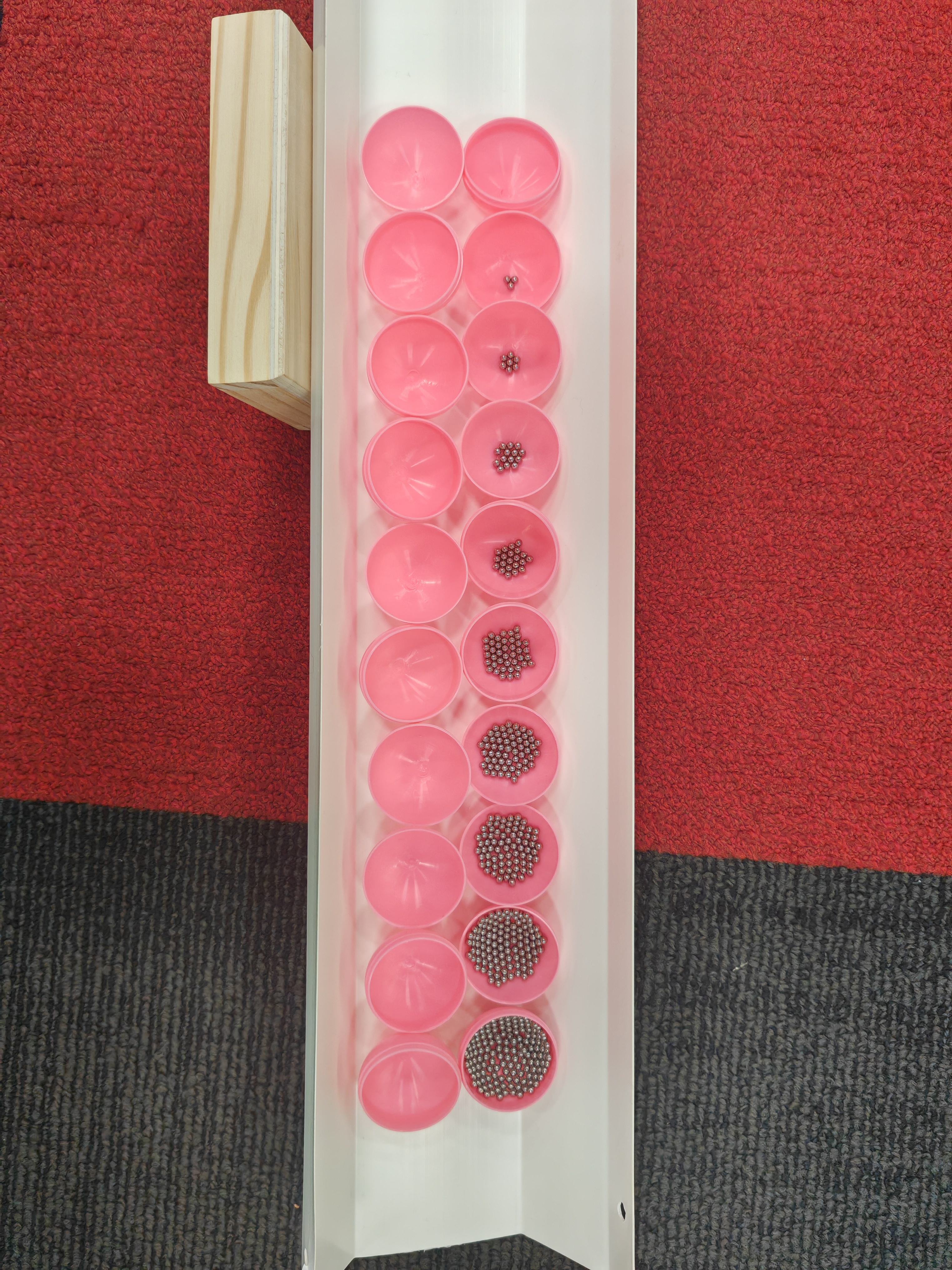}\\
{\footnotesize (a) Steel-bead loads for the ball variants}
\end{minipage}\hfill
\begin{minipage}[t]{0.48\linewidth}
\centering
\includegraphics[width=0.32\linewidth,height=0.16\textheight,keepaspectratio]{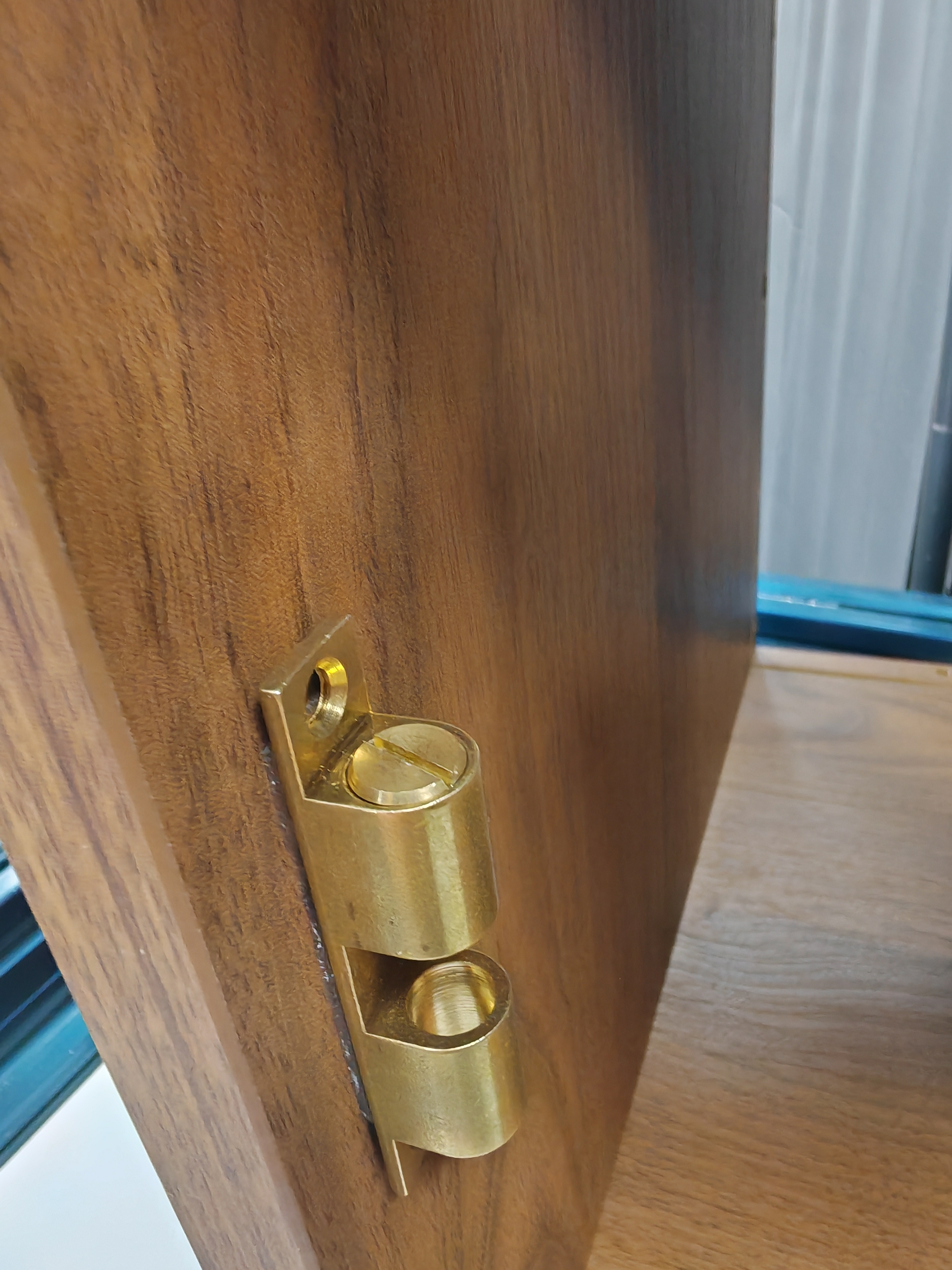}\hfill
\raisebox{0.05\textheight}{\includegraphics[width=0.65\linewidth,height=0.16\textheight,keepaspectratio,trim=0 300 0 200,clip]{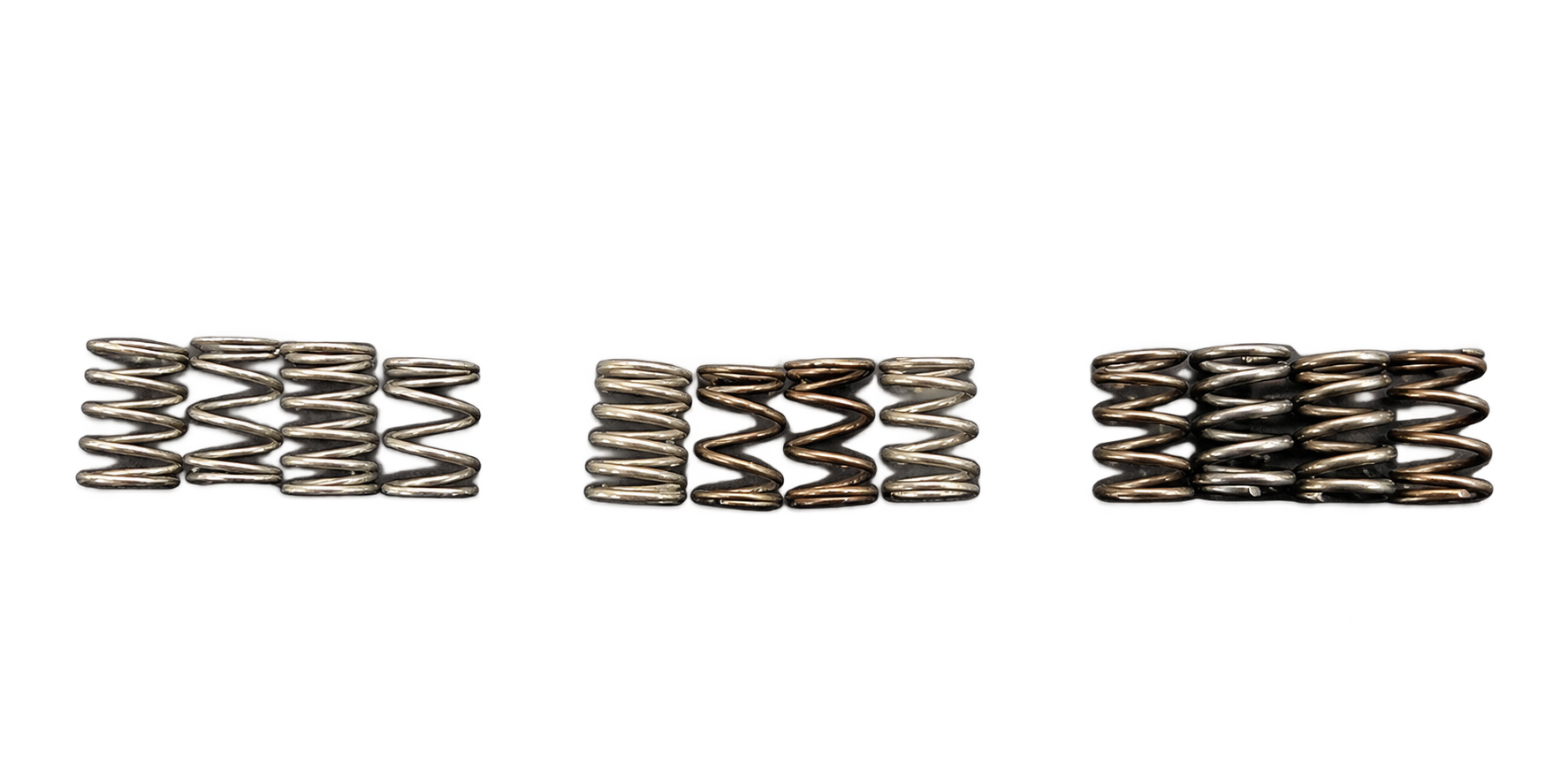}}\\
{\footnotesize (b) Door latch and alternative springs}
\end{minipage}
\vspace{0.6em}

\begin{minipage}[t]{0.48\linewidth}
\centering
\includegraphics[width=\linewidth,height=0.16\textheight,keepaspectratio]{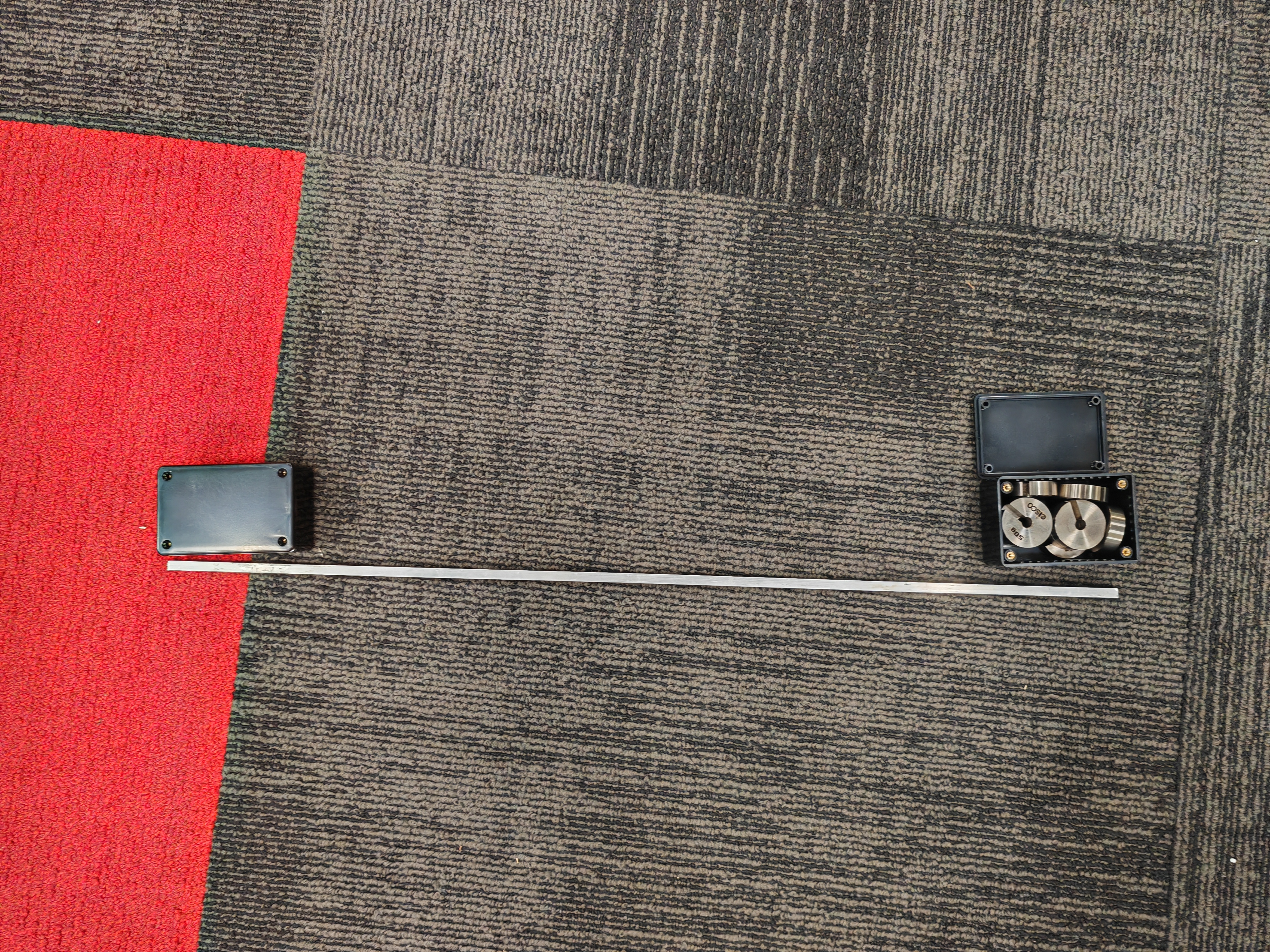}\\
{\footnotesize (c) Stick with adjustable end masses}
\end{minipage}\hfill
\begin{minipage}[t]{0.48\linewidth}
\centering
\includegraphics[width=\linewidth,height=0.16\textheight,keepaspectratio]{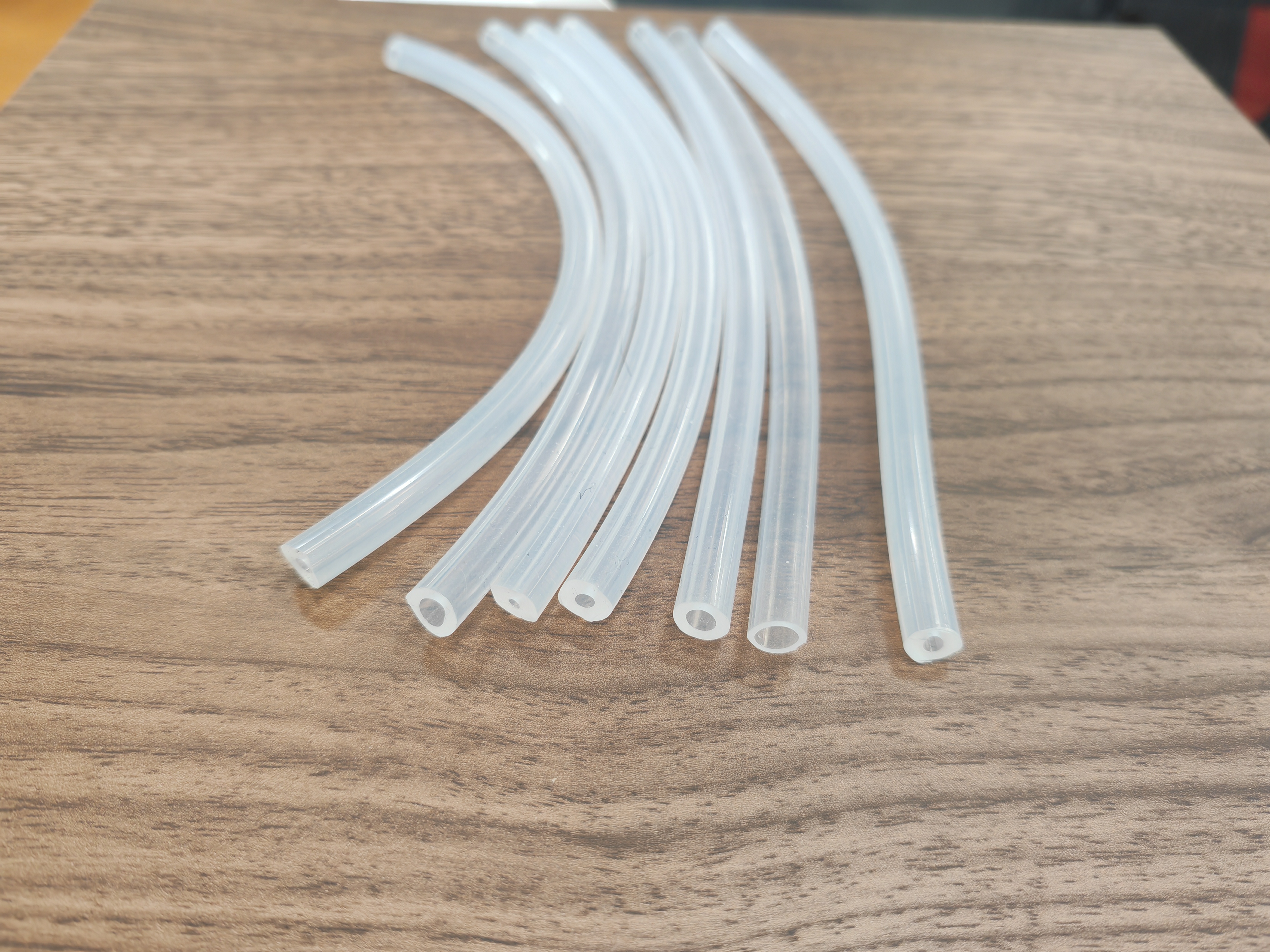}\\
{\footnotesize (d) Soft tubes with different stiffness and curvature}
\end{minipage}
\caption{Physical designs of the four real-world tasks. The Ball task changes
the number of steel beads inside the ball (not visible in the video frame),
and the Door task uses a spring-loaded latch.}
\label{fig:supp-real-designs}
\end{figure}

\paragraph{Ball Friction.}
We vary the mass of steel beads placed inside the ball, changing its total
mass and the sliding-friction force under a robot swing
(Figure~\ref{fig:supp-real-designs}a). The ball variants in the latent-space
visualizations correspond to these different steel-bead loads.
We track the ball's visible center for native-pixel ADE/FDE. Success means
that the selected swing moves the ball to the right-hand blue-marker target.

\paragraph{Door Close.}
The door has a spring-loaded latch; we replace the spring to change how much
closing action is needed to engage it (Figure~\ref{fig:supp-real-designs}b).
The tracked outer door edge, rather than a centroid or hinge angle, defines
native-pixel ADE/FDE. Success means that the selected action fully closes
and latches the door.

\paragraph{Stick Balance.}
We change the weights attached to the stick's ends, altering its mass
distribution and thus the
angle at which it settles after lifting (Figure~\ref{fig:supp-real-designs}c).
We track the two material markers at its ends and average their
positional errors for ADE/FDE. We sample a set of candidate actions and use
the model to predict which actions will maintain balance. One action is
selected uniformly at random from the predicted balancing set and executed
on the real robot. A trial is successful if the stick actually maintains
balance. Each sampled candidate set contains at least one action that
achieves balance. If the model predicts no balancing action, the trial is
counted as a failure. We report the fraction of successful trials.

\paragraph{Soft Pull.}
The task uses soft tubes with different stiffness coefficients. The robot
may grasp from the left or right, and the tubes also have different natural
curvatures (Figure~\ref{fig:supp-real-designs}d). The central prediction
question is approximately how far the grasp must rotate before the tube
starts dragging the object. The visible object center defines native-pixel
ADE/FDE. Given a candidate action, the model predicts the rotation angle
$\hat{\theta}$ at which the object starts sliding. We then conduct two
real-robot tests, rotating to $\hat{\theta}-6^\circ$ and
$\hat{\theta}+6^\circ$, respectively. A trial is successful only if the
object does not slide in the first test and slides in the second; otherwise,
it is counted as a failure. We report the fraction of successful trials.

\tzsection{Baseline Implementation Details}
\label{app:baselines}

All four baselines use the task-level training data and per-method GPU
budgets in Tables~\ref{tab:supp-sim-train}--\ref{tab:supp-real-train}.
They use the same query videos and, for action-selection tasks, the same
candidate actions and success criteria as T2-Z. The three adaptive
baselines receive exactly the same test-time interaction trajectories as our latent
optimizer.
The tables give the common task-level batch specification; DINOv2 Context
uses a subset of sampled trajectories for conditioning and all sampled trajectories
as queries during training, whereas TTT-KQV
processes them as ordered within-environment streams.

\tzsubsection{Standard World Model}
This baseline fine-tunes the BWM-initialized DiT and action encoder on
pooled task trajectories without an environment latent or test-time
adaptation. It uses AdamW with model learning rate $10^{-5}$, weight decay
$0.01$, 100 warmup updates, gradient clipping $0.5$, and bfloat16.

\tzsubsection{LoRA TTA}
Starting from the Standard World Model, rank-8, $\alpha=8$ LoRA
adapters~\citep{lora}
are placed on the DiT attention Q/K/V/output projections. The backbone
stays frozen during test-time adaptation. Adapters are reset for each
environment and optimized on its test-time interaction trajectories, then held fixed
across that environment's query actions. The Friction configuration uses 50
AdamW adaptation updates at learning rate $10^{-4}$, zero weight decay, and
gradient clipping $1$; task-specific adaptation configurations are
released with the code.

\tzsubsection{DINOv2 Context}
A frozen DINOv2-B/14 encoder~\citep{dinov2} extracts conditioning-frame features, while a
learned 128-D action GRU summarizes the intervening 14-D actions. For each
of $N$ task-configured sampled frames, the CLS and mean patch features are
projected to 256 dimensions. The $N-1$ adjacent visual--action transition
features (640-D each) are concatenated; a $640(N-1)\!\to\!1024\!\to\!1024$
MLP and final projection produce the 32-D context. Multiple conditioning-trajectory
summaries are averaged before that final projection.
At each training update, we sample a set of trajectories $Q_e$ from a training
environment and select a subset $C_e\subseteq Q_e$ for conditioning. All
trajectories in $Q_e$ serve as training queries. The number of conditioning trajectories is sampled from the
task-configured choices (typically one or two; one to four for Soft Pull),
so the predictor learns to use different conditioning-set sizes. Every query
in $Q_e$ is predicted using only the context inferred from $C_e$; training
minimizes the mean query-video flow-matching loss. The context
head, BWM DiT, action encoder, and two context projectors are trained
jointly, while DINOv2 remains frozen. The model and context-head learning
rates are $10^{-5}$ and $10^{-4}$, respectively. At test time, context
comes from a forward pass over the observed test-time interaction trajectories, with no gradient
update.

\tzsubsection{TTT-KQV}
We adapt a video test-time-training fast-weight module~\citep{dalal2025oneminute}
to the BWM backbone. Gated two-layer GELU MLP modules with expansion
factor four are inserted after eight approximately evenly spaced
attention blocks. Training randomly shuffles the chunks from each sampled
environment in a batch. Within each training stream, all chunks use the same
sampled flow-matching timestep (but independent noise), matching the
same-timestep processing of observed interactions and query videos at inference.
At position $i$, the preceding chunks form the observed prefix and the current chunk is the query. Fast weights
reset to a learned shared initialization at the start of each stream.
For each chunk, the module causally writes its key--value tokens before
reading its output. Let $x_t$ denote an input token, $h$ an attention head,
and $d_h$ its feature dimension; $t$ indexes tokens, not flow-matching
timesteps. Learned key and value projections produce $k_{h,t},v_{h,t}
\in\mathbb R^{d_h}$. The fast state $W=\{W_h\}_h$ contains each head's
MLP weights and biases. Its per-token binding loss and learned step scaling are
\begin{equation}
\begin{aligned}
\ell_{h,t}(W)
&=\frac{1}{2d_h}\left\|k_{h,t}
 +\mathrm{LN}\!\left(\mathrm{MLP}_{W_h}(k_{h,t})\right)-v_{h,t}\right\|_2^2,\\
\eta_{h,t}
&=\frac{\eta_{\mathrm{base}}}{d_h}\,
 \sigma(w_h^\top x_t+b_h),\qquad \eta_{\mathrm{base}}=0.1.
\end{aligned}
\label{eq:supp-ttt-binding}
\end{equation}
Here $\mathrm{LN}$ normalizes within each head, and $\sigma$ is the sigmoid
function. The factor $1/d_h$ in the loss averages squared errors over feature
dimensions; the separate factor $1/d_h$ in $\eta_{h,t}$ scales the update.
For update $j$, let $\mathcal B_j$ be the current minibatch of at most 64 tokens.
The differentiable fast-state update is
\begin{equation}
W^{j+1}=W^j-
\left.\nabla_W\sum_h\sum_{t\in\mathcal B_j}
\eta_{h,t}\ell_{h,t}(W)\right|_{W=W^j}.
\label{eq:supp-ttt-update}
\end{equation}
We use a base inner learning rate of $0.1$ with learned token-wise,
head-specific scaling, rather than a constant update step size.
The scaling parameters $w_h,b_h$ are shared slow weights; the inner update
changes only $W$. Gates start at $0.1$. The outer loss averages the video
flow-matching losses over all stream positions and sampled environments.
Its gradients update the shared slow weights, using backbone LR $10^{-5}$
and TTT-module LR $10^{-4}$; the fast state itself is ephemeral.
At inference, every denoising timestep resets the fast state and replays
the observed test-time interaction trajectories at that same timestep. Each query then takes an
independent update--then--read branch on its own noisy tokens. Query
ground truth never enters the update, and fast state is not carried
between queries or denoising timesteps. Exact stream packing and
task-specific execution settings are released with the code.

\clearpage
\tzsection{Additional Experiments}
\tzsubsection{Test-time interaction-count ablation}
\begin{table}[H]
\centering\small
\caption{Numerical results for Figure~5(c,d) at $K=1,2,4$ test-time
interaction trajectories. Bold marks the selected $K$. Physical error is in pixels
except for Light Switch (lamp-score error in percentage points).}
\label{tab:supp-support-count}
\setlength{\tabcolsep}{4pt}
\begin{tabular*}{\linewidth}{@{\extracolsep{\fill}}lcccccc@{}}
\toprule
& \multicolumn{3}{c}{Physical error (mean) $\downarrow$}
& \multicolumn{3}{c}{Action success (\%) $\uparrow$} \\
\cmidrule(lr){2-4}\cmidrule(lr){5-7}
Task & $K=1$ & $K=2$ & $K=4$ & $K=1$ & $K=2$ & $K=4$ \\
\midrule
Mass Collision & \textbf{6.452} & 6.534 & 5.946
& \textbf{61.1} & 66.7 & 72.2 \\
Light Switch & 8.26 & \textbf{3.43} & 3.41
& 50.0 & \textbf{87.5} & 87.5 \\
Door Close & 11.337 & \textbf{7.954} & 6.889
& 22.2 & \textbf{55.6} & 55.6 \\
Soft Pull & \textbf{18.710} & 18.363 & 18.060
& \textbf{81.8} & 81.8 & 81.8 \\
\bottomrule
\end{tabular*}
\end{table}

\tzsubsection{Latent dimensionality}
\begin{table}[H]
\centering\footnotesize
\caption{Environment-latent dimensionality on Friction under the same
$K=1$ evaluation protocol (ten held-out physical environments;
90 query videos). Object ADE/FDE are main-view
centroid errors in pixels; video metrics use the concatenated main and wrist
views. Action success is evaluated on 25 reachable target decisions. Best
values are bold.}
\label{tab:supp-latent-dim}
\setlength{\tabcolsep}{3pt}
\begin{tabular*}{\linewidth}{@{\extracolsep{\fill}}lrrrrrr@{}}
\toprule
Latent dim. & ADE$\downarrow$ & FDE$\downarrow$ & PSNR$\uparrow$ & SSIM$\uparrow$ & LPIPS$\downarrow$ & Success$\uparrow$ \\
\midrule
$8$ & 8.92 & 19.26 & 31.293 & 0.9474 & 0.0646 & 44\% \\
$16$ & 7.63 & 15.90 & 29.557 & 0.9470 & 0.0614 & 64\% \\
$32$ (ours) & \textbf{3.85} & \textbf{7.59} & \textbf{31.976} & \textbf{0.9513} & \textbf{0.0495} & \textbf{72\%} \\
$64$ & 6.86 & 13.98 & 31.692 & 0.9504 & 0.0551 & 64\% \\
$128$ & 7.08 & 14.51 & 31.372 & 0.9483 & 0.0581 & 64\% \\
\bottomrule
\end{tabular*}
\end{table}

\begin{figure}[p]
\centering
\includegraphics[width=0.8\linewidth]{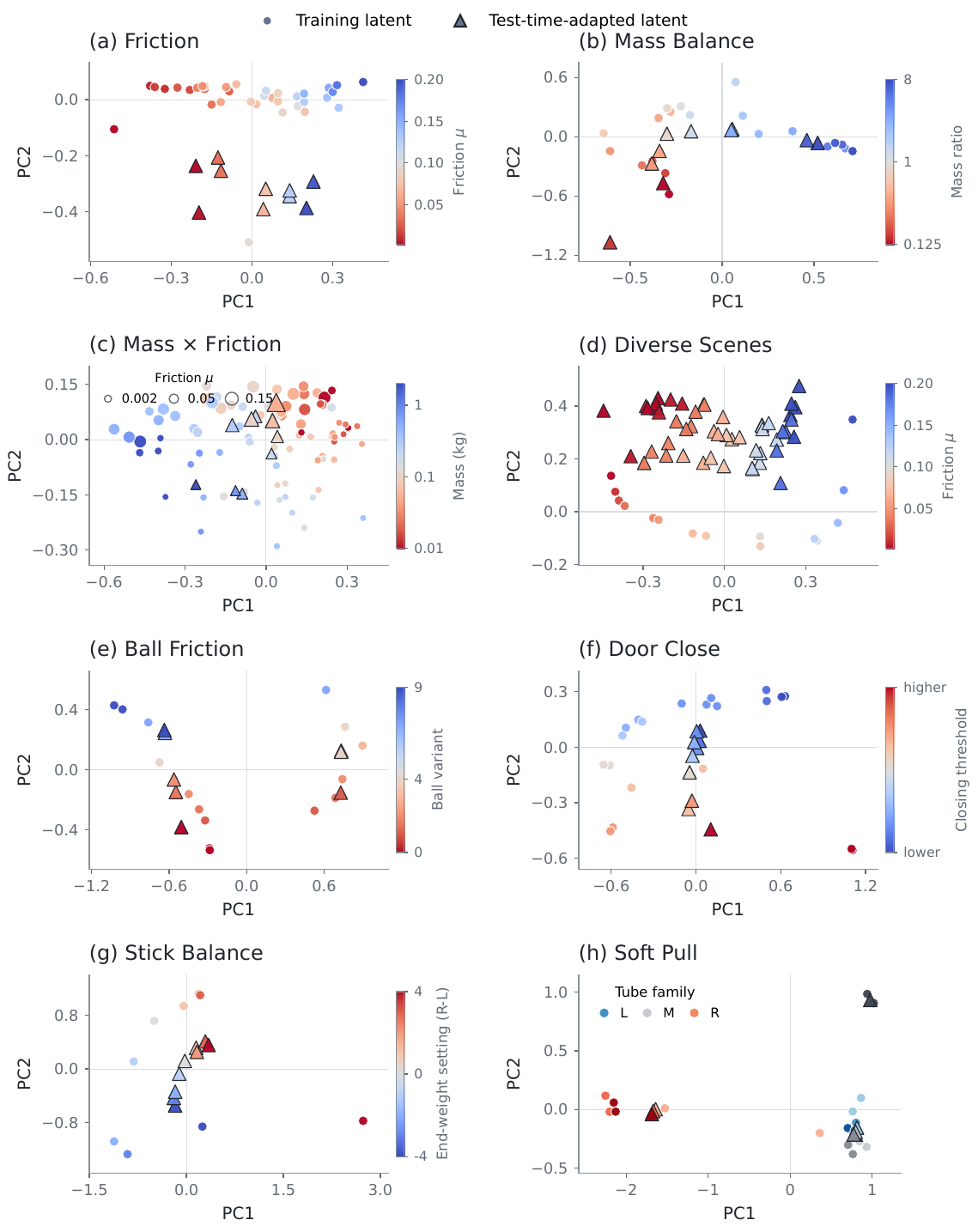}
\caption{Latent-space PCA across simulation (a--d) and real-world tasks
(e--h). Circles denote training-time latents and triangles denote
test-time-adapted latents. Color indicates the underlying physical setting;
point size additionally indicates friction in Mass $\times$ Friction. These
factor annotations are used only for visualization and are not provided
to the model. The Soft Pull test-time latents shown here use four test-time interaction
trajectories.}
\label{fig:supp-pca}
\end{figure}

\tzsubsection{Latent-space PCA across tasks}
\label{app:pca}
Figure~\ref{fig:supp-pca} shows the latent-space analysis for four
simulation and four real-world tasks.
Circles show training-time environment latents, and triangles show latents
adapted from test-time interaction trajectories. Colors indicate the underlying
physical setting or causal class for visualization only; these labels are
not provided to the model. For each task independently, we center its
training-time 32-D latents by their mean $\bar z$, compute the leading two
right singular vectors $V_{1:2}$ of the centered matrix without feature
rescaling, and plot $(z-\bar z)V_{1:2}$. Test-time latents are projected
using the same mean and basis; they do not enter the PCA fit. Thus the PC
axes are task-specific rather than shared across panels.

\makeatletter\setlength{\@fptop}{0pt}\makeatother
\makeatletter\setlength{\@fptop}{0pt plus 1fil}\makeatother

\begin{figure}[htbp]
\centering
\includegraphics[width=0.6\linewidth]{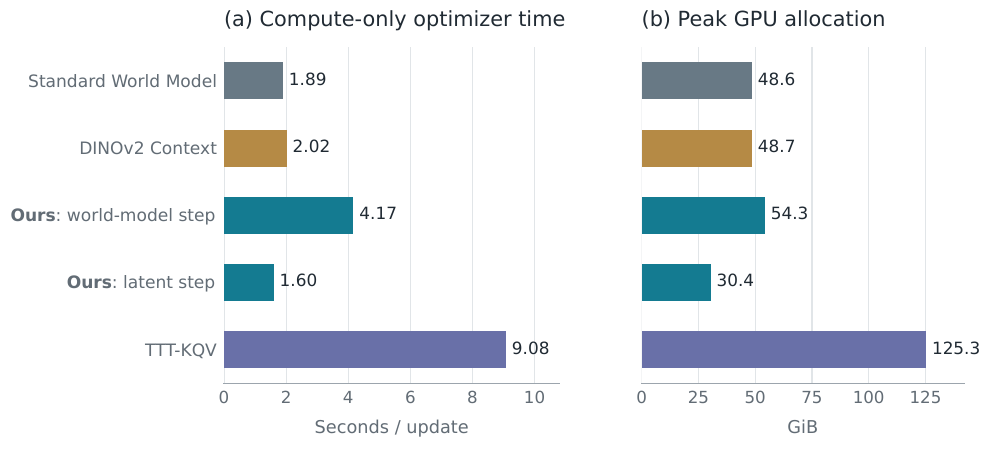}
\caption{Training resource profile. Time is the median compute-only optimizer
duration over ten measured updates after two warmup updates, excluding dataset
reads; memory is peak allocated GPU memory. Each update uses six loss-bearing
video chunks.}
\label{fig:supp-efficiency-training}
\end{figure}

\tzsubsection{Computational efficiency}
\label{app:efficiency}
We profile a representative Friction setting on one H200 GPU.
For training, each method is warmed up for two optimizer steps and timed for
ten further steps. Figure~\ref{fig:supp-efficiency-training} reports the
median compute-only optimizer time per update, excluding dataset reads, and
peak allocated GPU memory rather than full-run training time. Each measured
update has six loss-bearing video chunks; method-specific conditioning
computation remains included. The world-model and latent updates of our method
are separate phases and are therefore shown separately. LoRA TTA starts from
the Standard World Model and has no distinct world-model training row.

\begin{figure}[H]
\centering
\includegraphics[width=\linewidth]{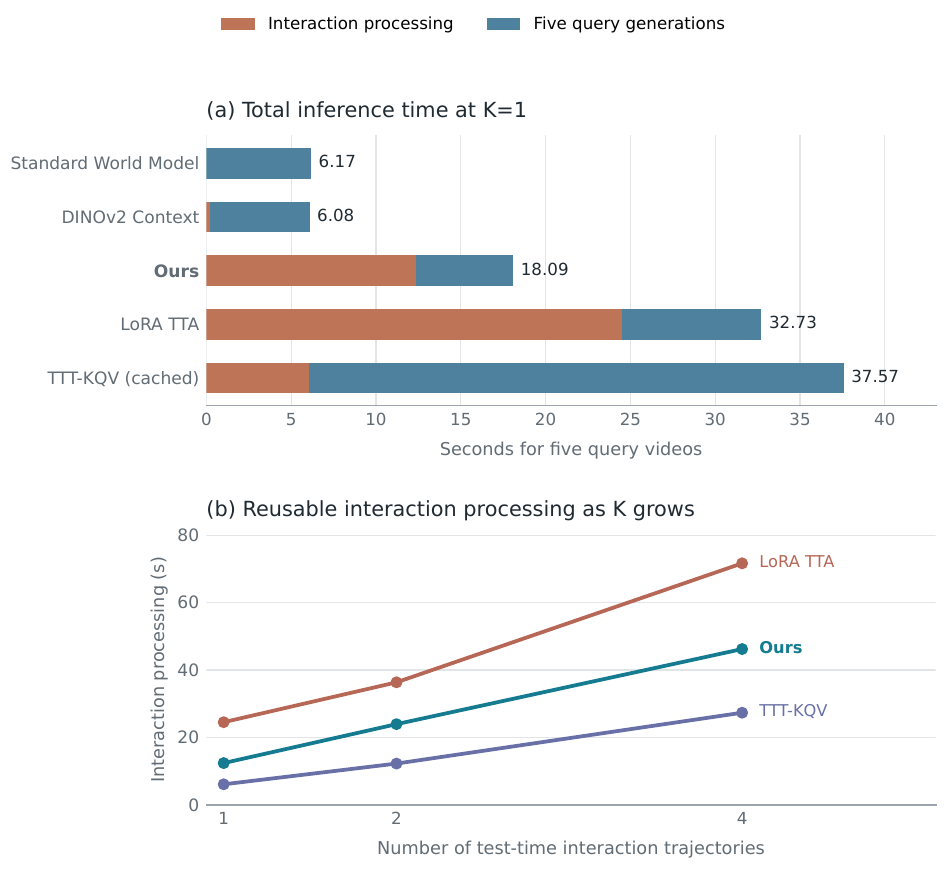}
\caption{Compute-only inference profile on one H200 GPU, excluding model and
sample loading and MP4 writing. Top: processing of one test-time interaction
trajectory plus generation of the same five query videos. Bottom: interaction
processing time for $K=1,2,4$ with the update counts held fixed within each
method. The bottom panel does not compare query-generation time at $K>1$.}
\label{fig:supp-efficiency-inference}
\end{figure}

For inference, all methods receive the same test-time interaction trajectory and five
disjoint query video chunks. We use 25 denoising steps and exclude model
loading, dataset reads, and MP4 writing from timing; query generation includes
pipeline sampling and VAE decoding. Interaction processing is performed once
and reused across queries.
DINOv2 Context encodes the interaction trajectory without gradient updates, our method runs
40 latent updates, and LoRA TTA runs 50 adapter updates. TTT-KQV precomputes
one post-interaction fast state at each denoising timestep and reuses those
states for all five queries. Figure~\ref{fig:supp-efficiency-inference}
separates interaction processing from five-query generation and shows how processing
time changes with $K\in\{1,2,4\}$; the five-method total-time comparison
uses $K=1$ throughout. At $K=1$, our method takes 18.09 s including latent
adaptation and five query generations, versus 32.73 s for LoRA TTA and
37.57 s for cached TTT-KQV.

\clearpage

\begin{figure}[p]
\centering\includegraphics[width=.90\linewidth]{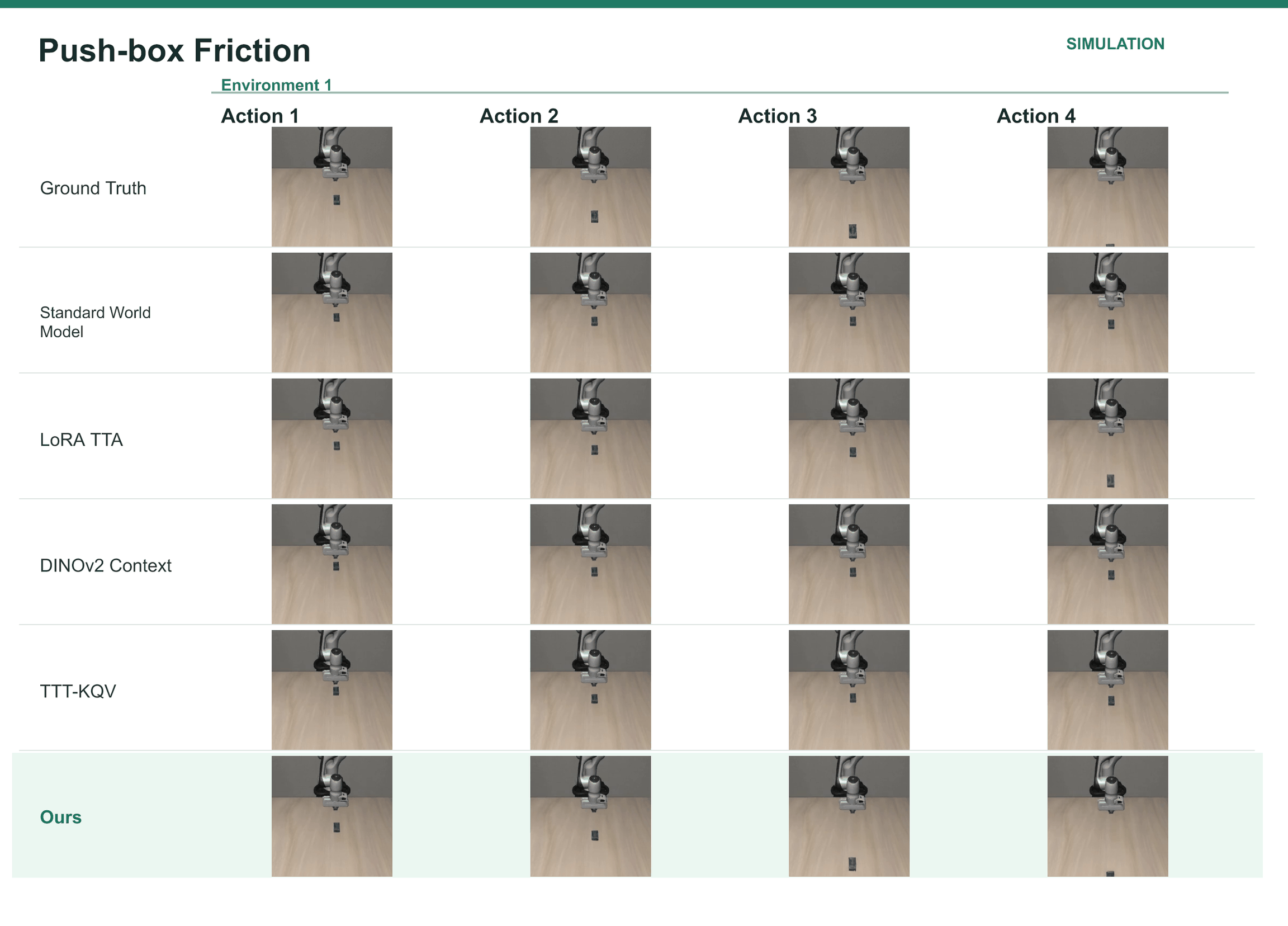}\par
\includegraphics[width=.90\linewidth]{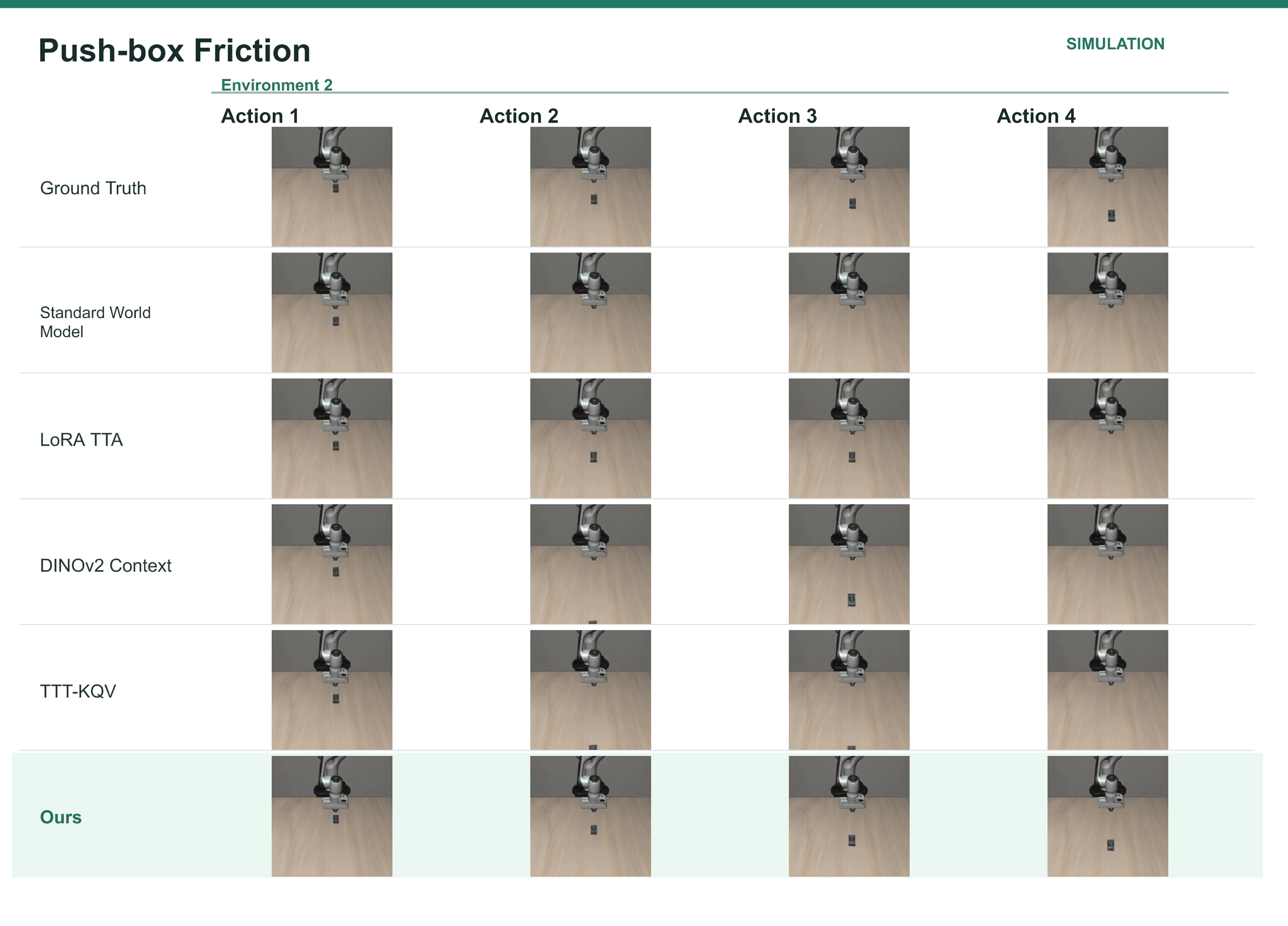}
\caption{Friction: selected test actions.}\label{fig:qual-friction}
\end{figure}

\begin{figure}[p]
\centering\includegraphics[width=.90\linewidth]{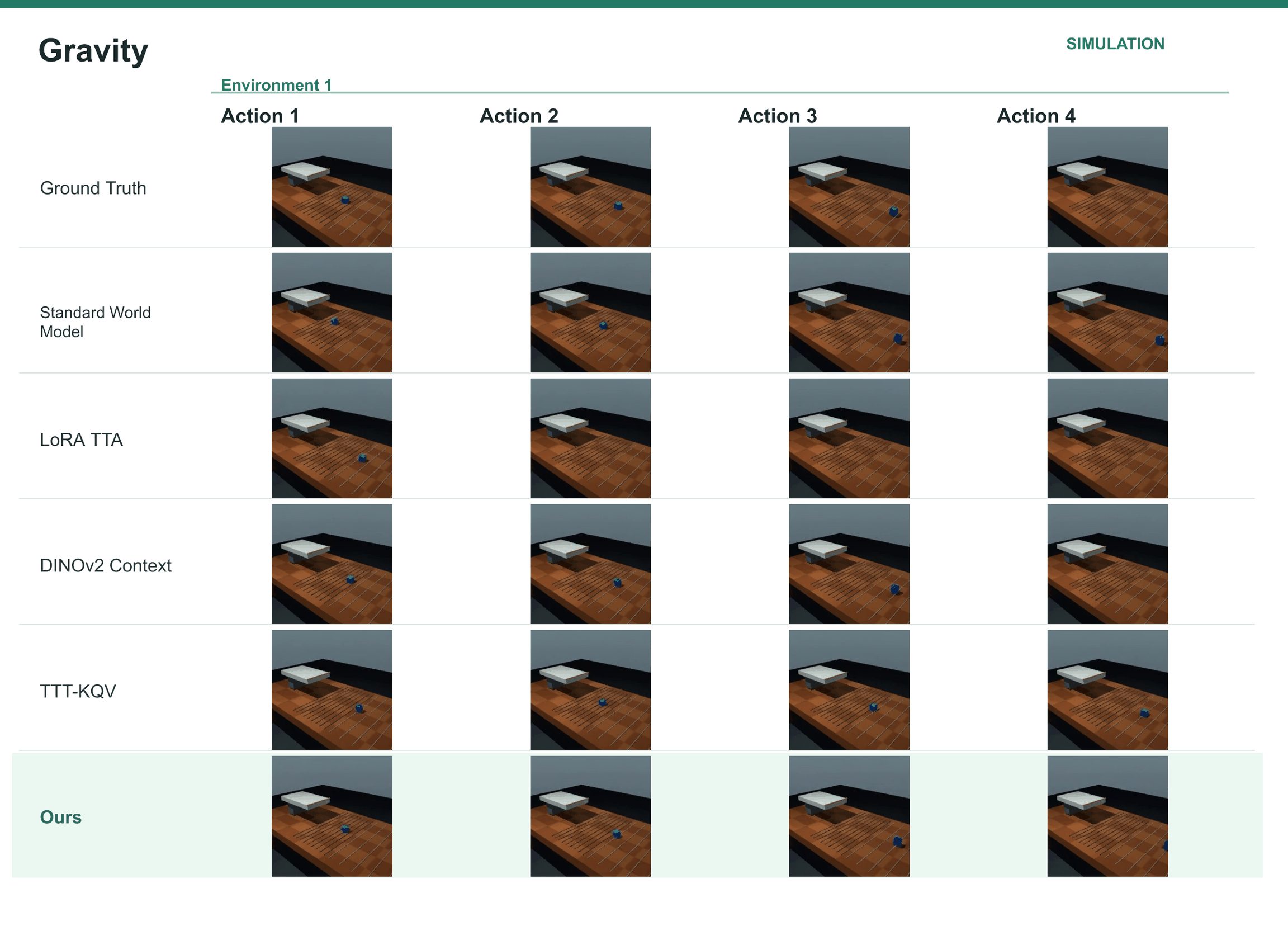}\par
\includegraphics[width=.90\linewidth]{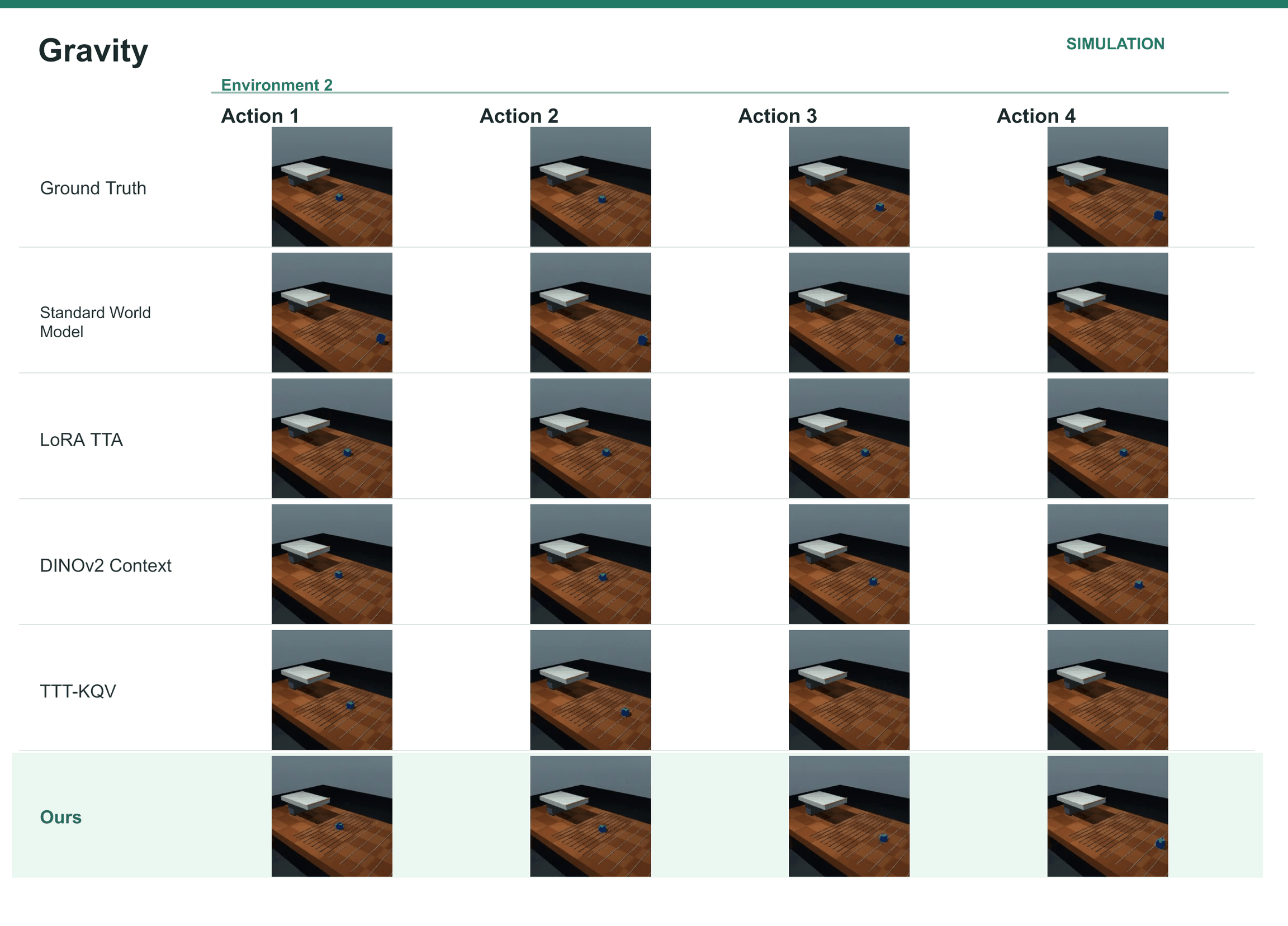}
\caption{Gravity: selected test actions.}\label{fig:qual-gravity}
\end{figure}

\begin{figure}[p]
\centering\includegraphics[width=.90\linewidth]{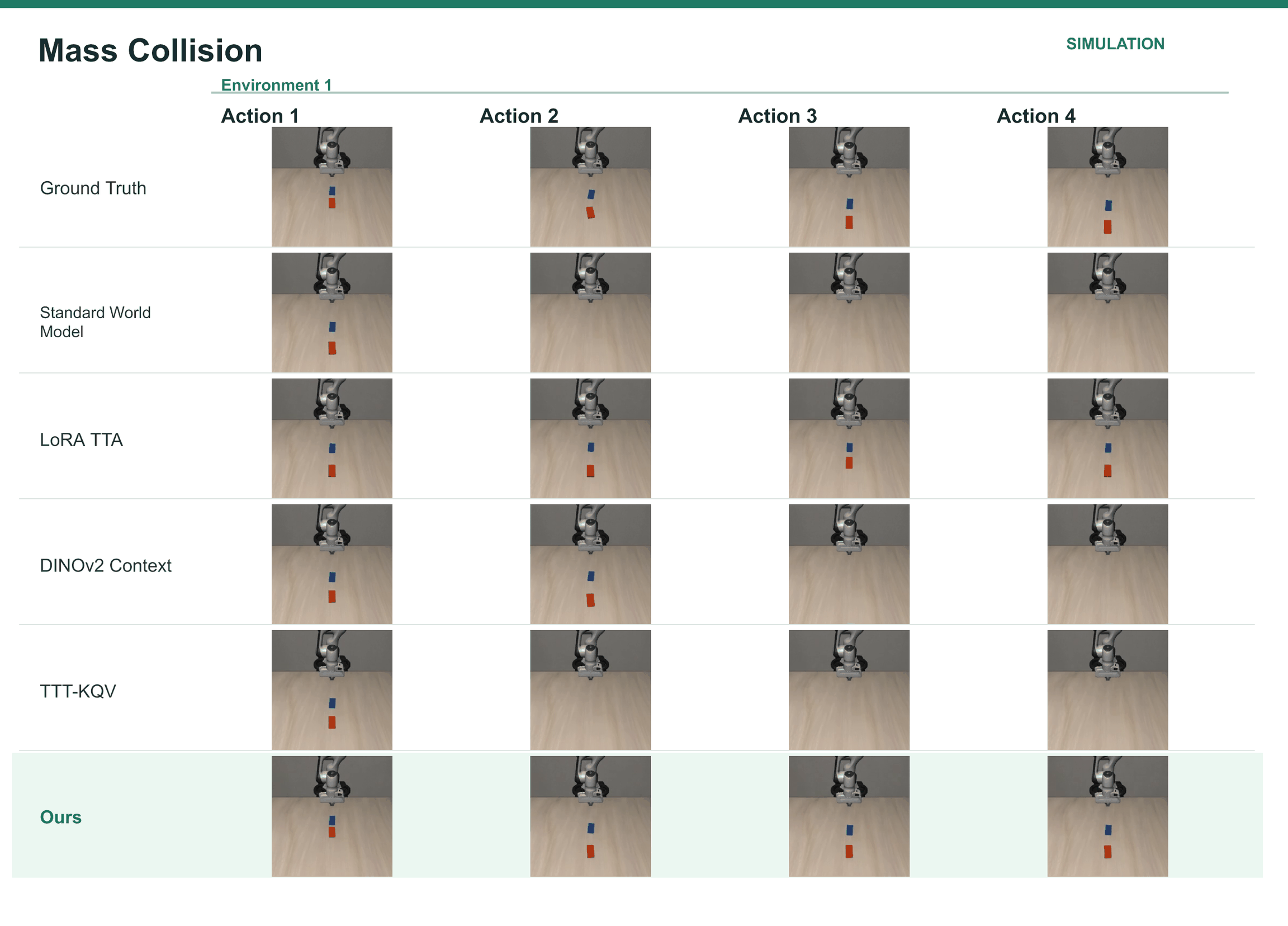}\par
\includegraphics[width=.90\linewidth]{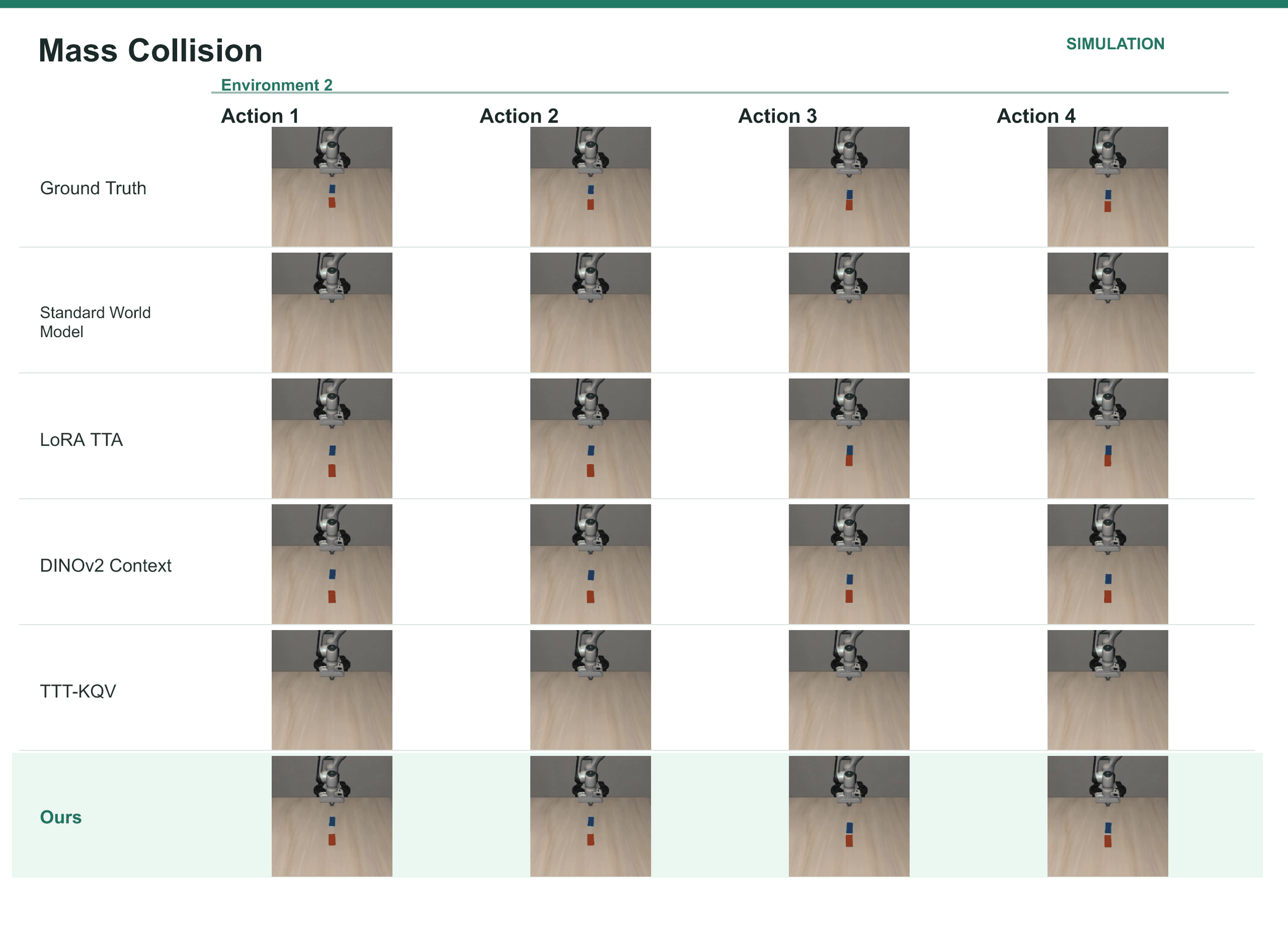}
\caption{Mass Collision: selected test actions.}\label{fig:qual-collision}
\end{figure}

\begin{figure}[p]
\centering\includegraphics[width=.90\linewidth]{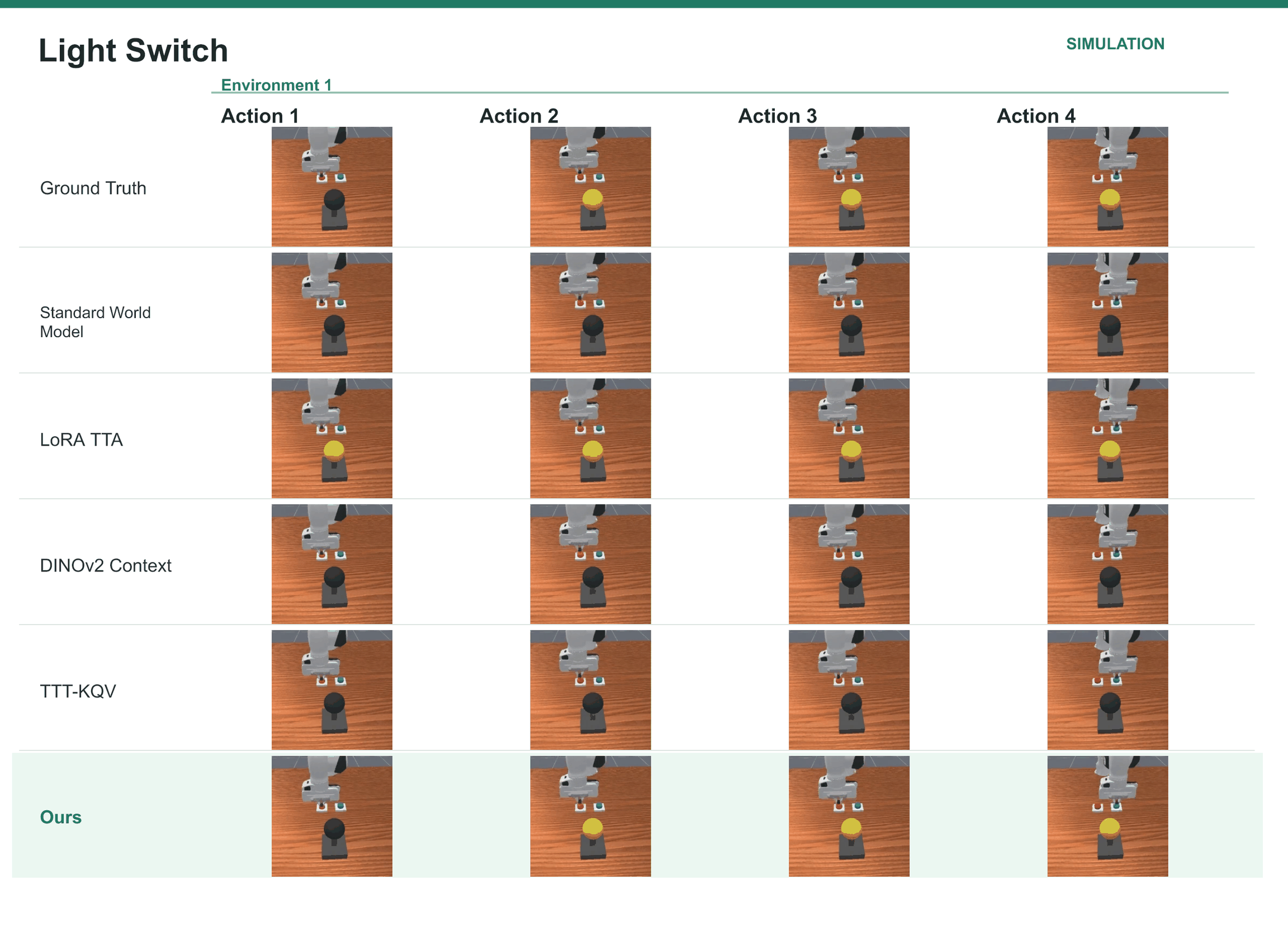}\par
\includegraphics[width=.90\linewidth]{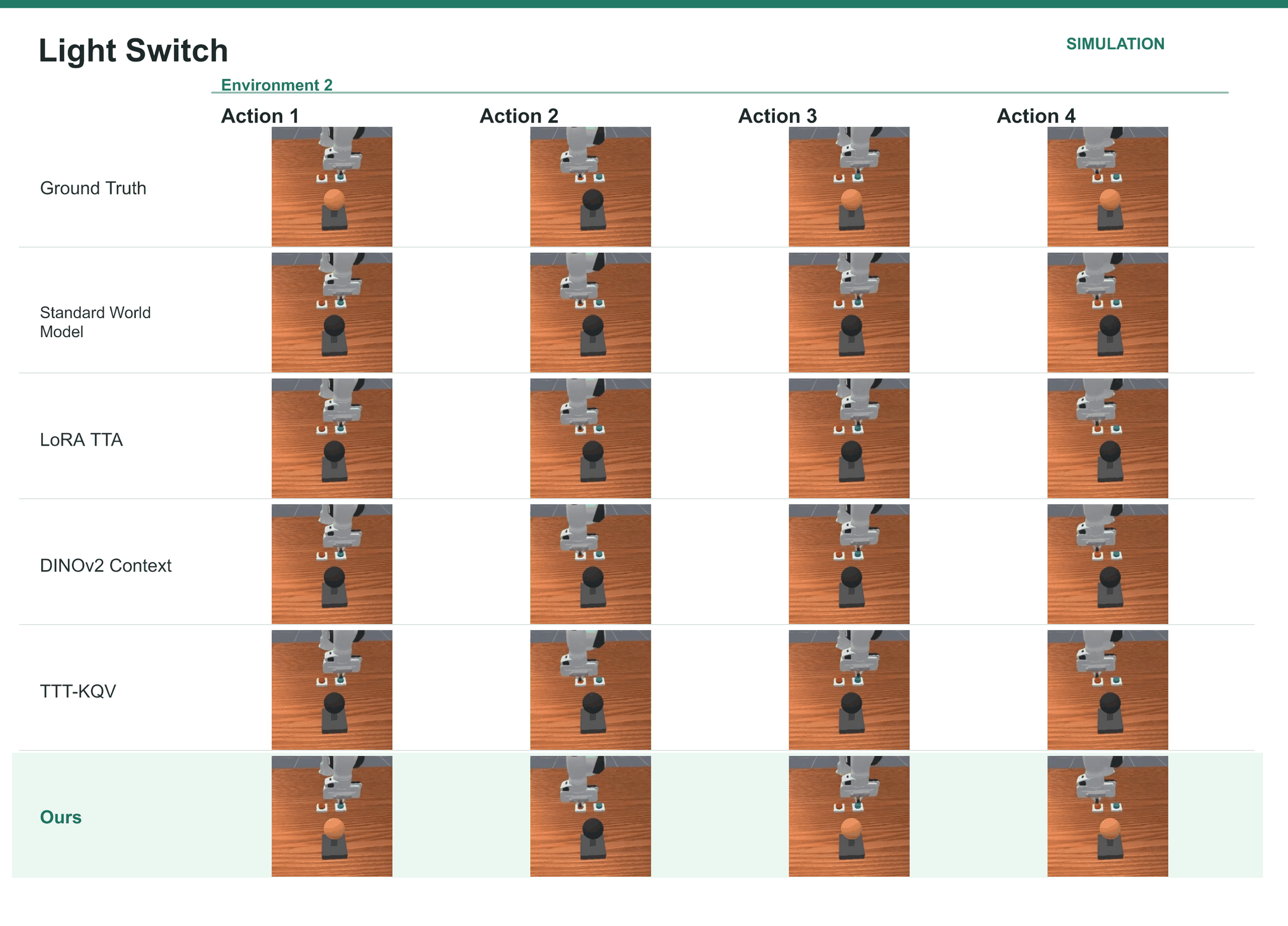}
\caption{Light Switch: selected test actions.}\label{fig:qual-light}
\end{figure}

\begin{figure}[p]
\centering\includegraphics[width=.90\linewidth]{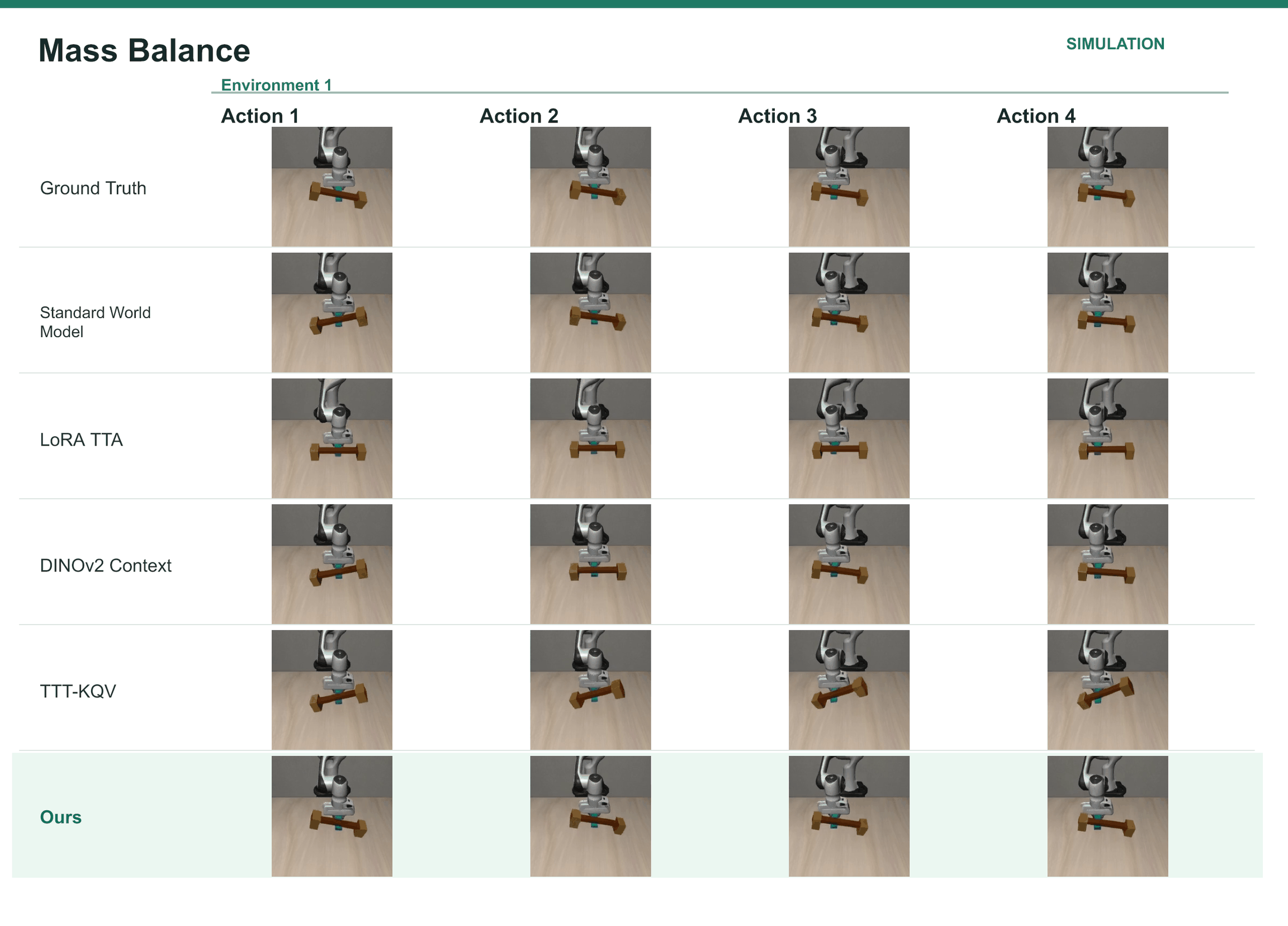}\par
\includegraphics[width=.90\linewidth]{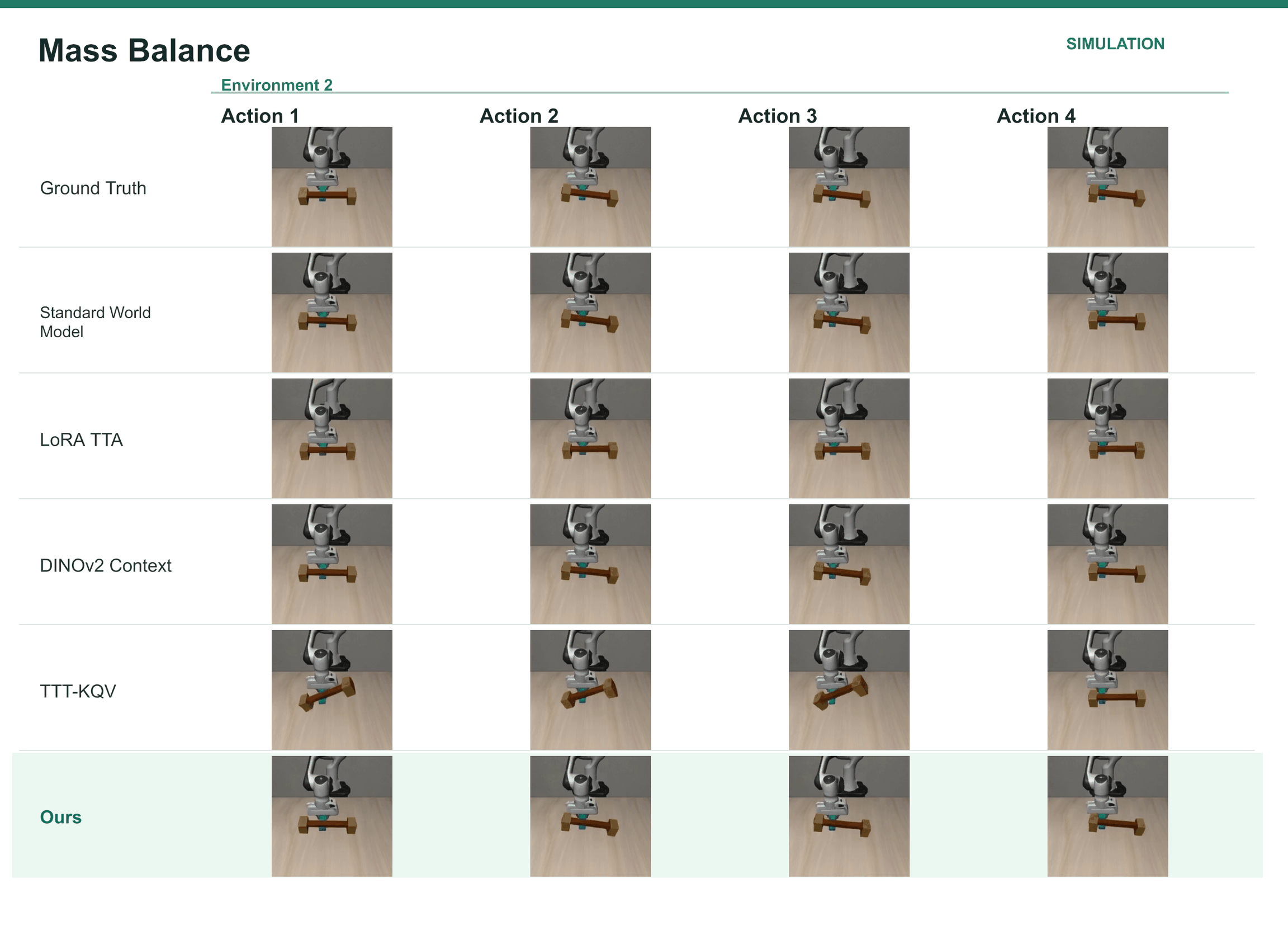}
\caption{Mass Balance: selected test actions.}\label{fig:qual-balance}
\end{figure}

\begin{figure}[p]
\centering\includegraphics[width=.90\linewidth]{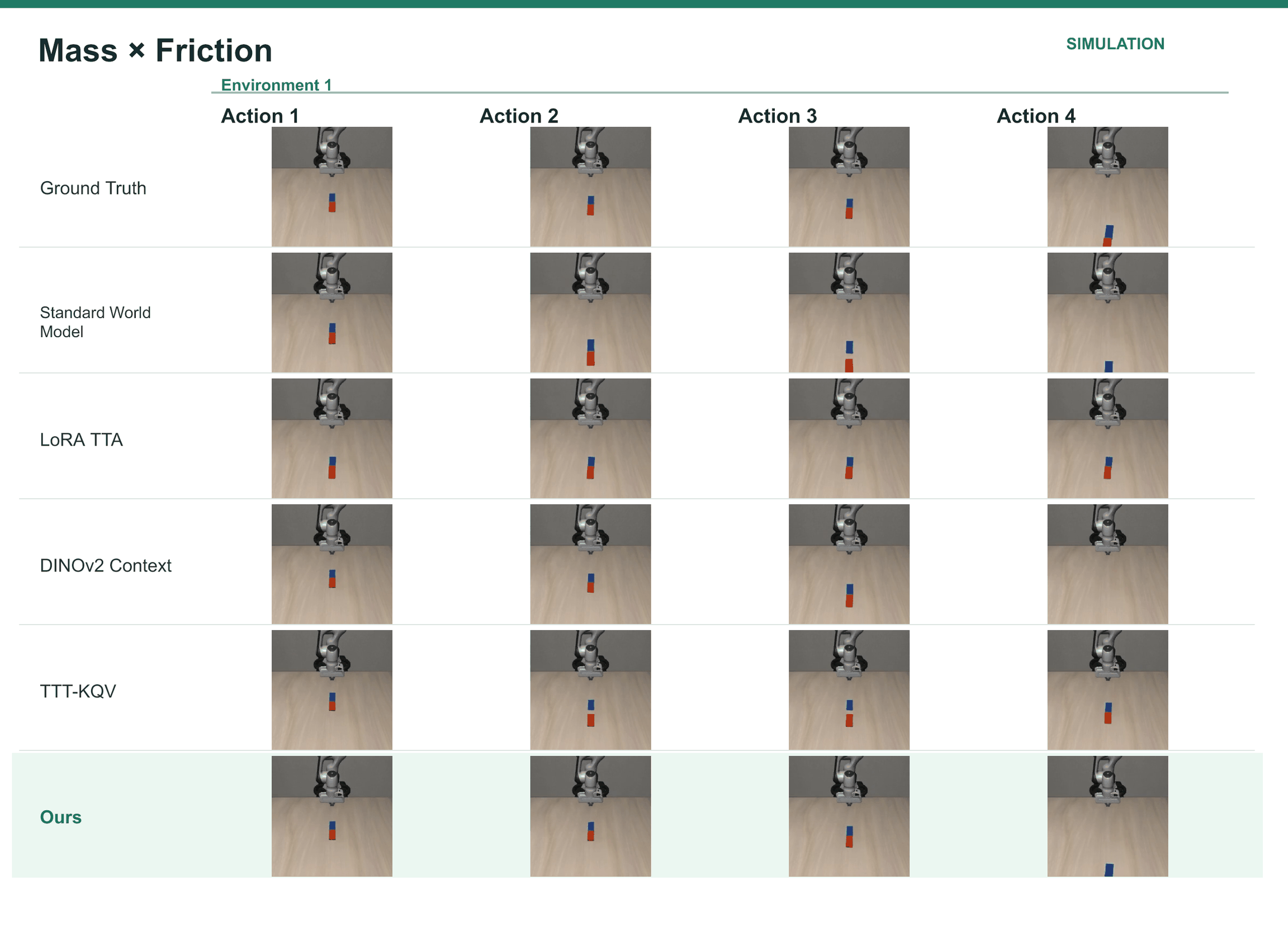}\par
\includegraphics[width=.90\linewidth]{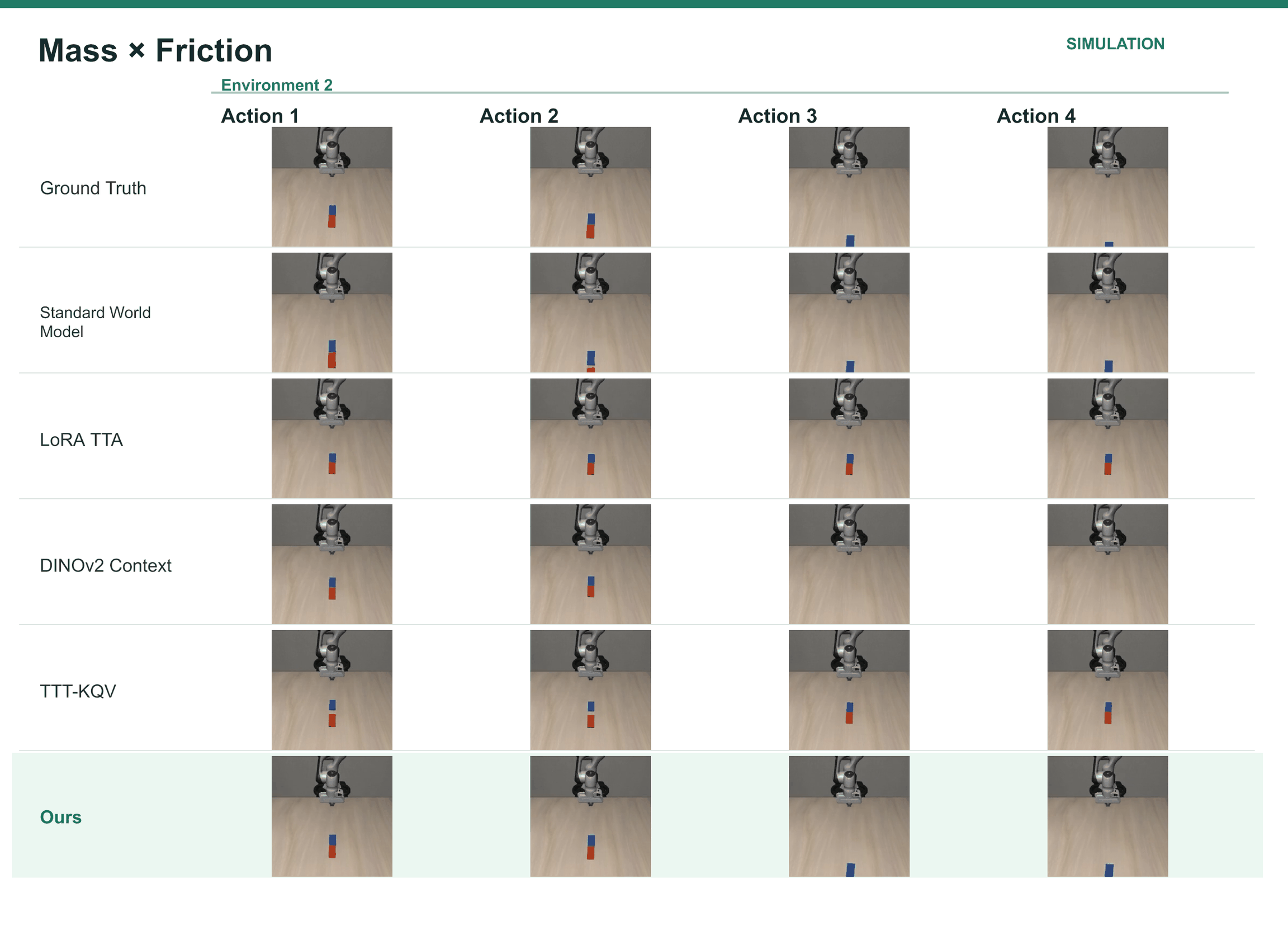}
\caption{Mass $\times$ Friction: selected test actions.}\label{fig:qual-mass-friction}
\end{figure}

\begin{figure}[p]
\centering\includegraphics[width=.90\linewidth]{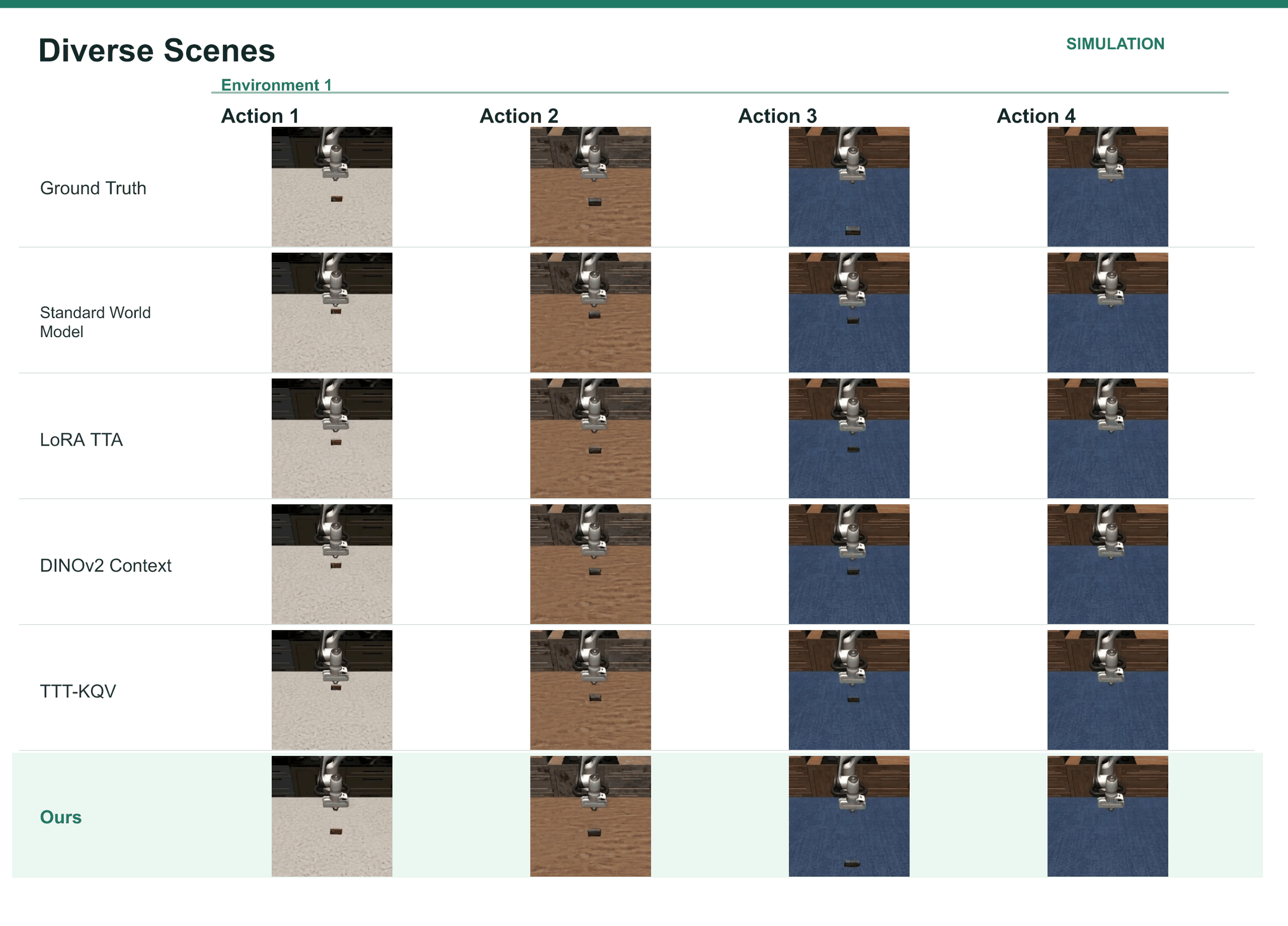}\par
\includegraphics[width=.90\linewidth]{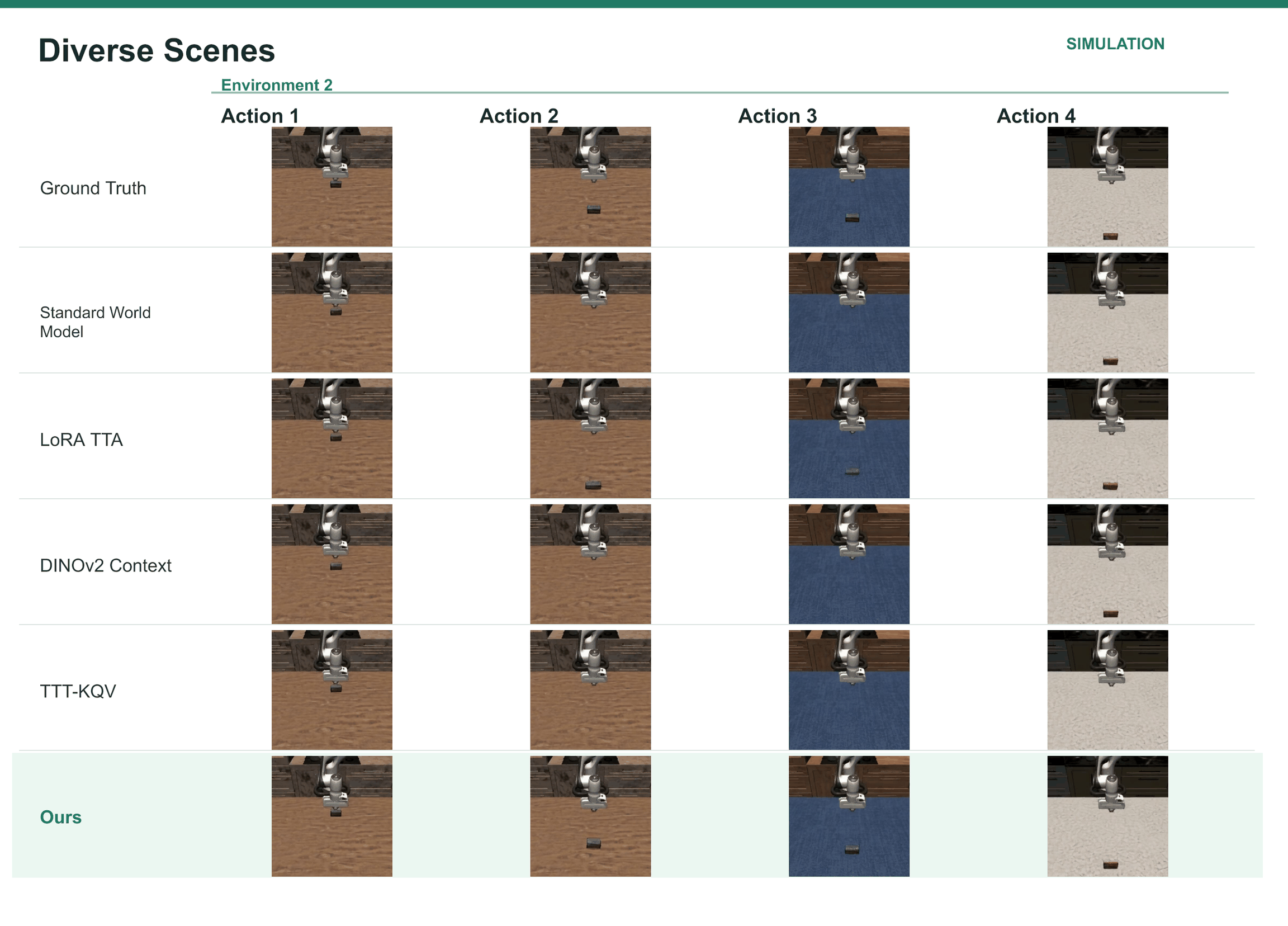}
\caption{Diverse Backgrounds: selected test actions.}\label{fig:qual-diverse}
\end{figure}

\begin{figure}[p]
\centering\includegraphics[width=.90\linewidth]{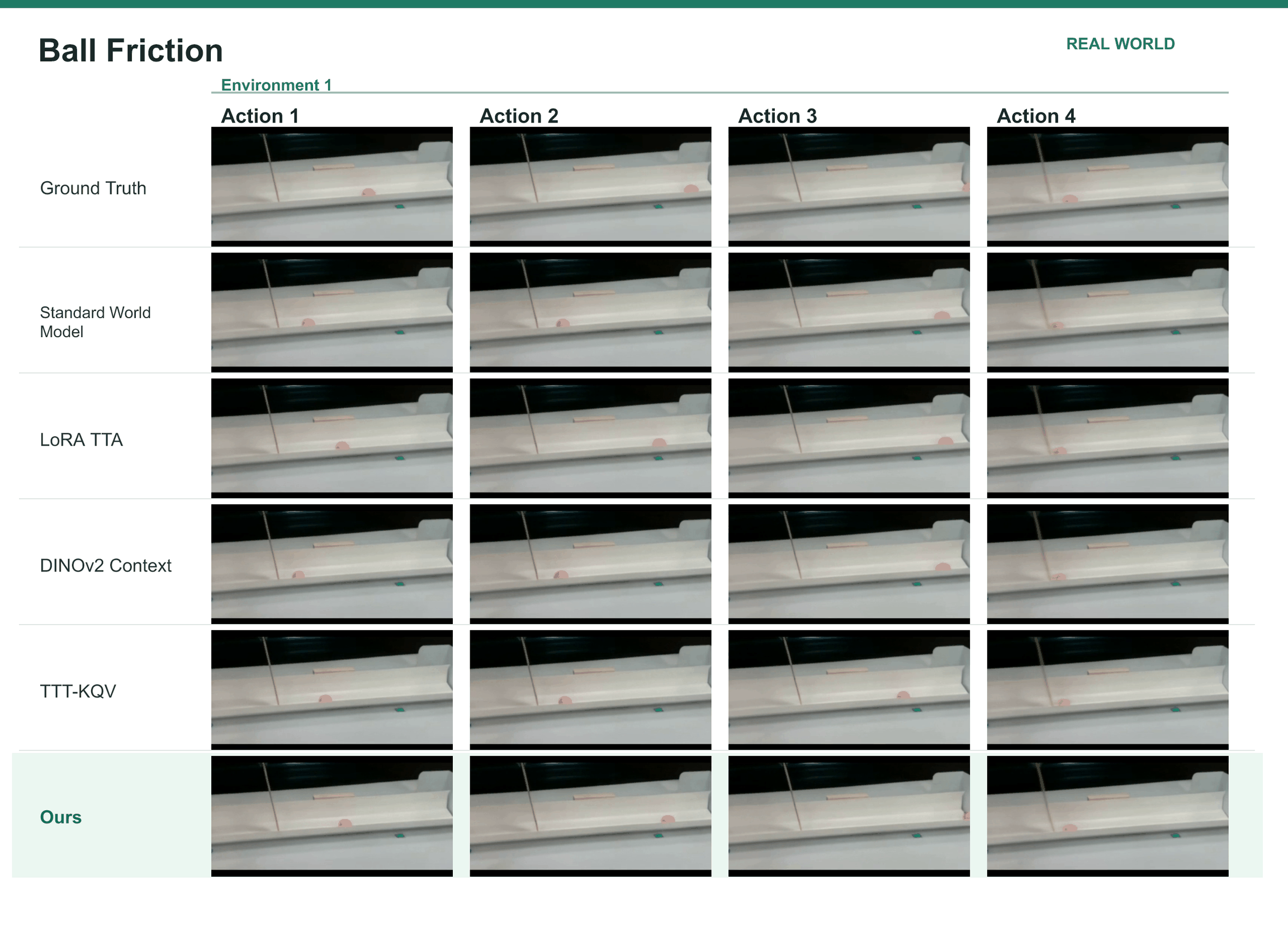}\par
\includegraphics[width=.90\linewidth]{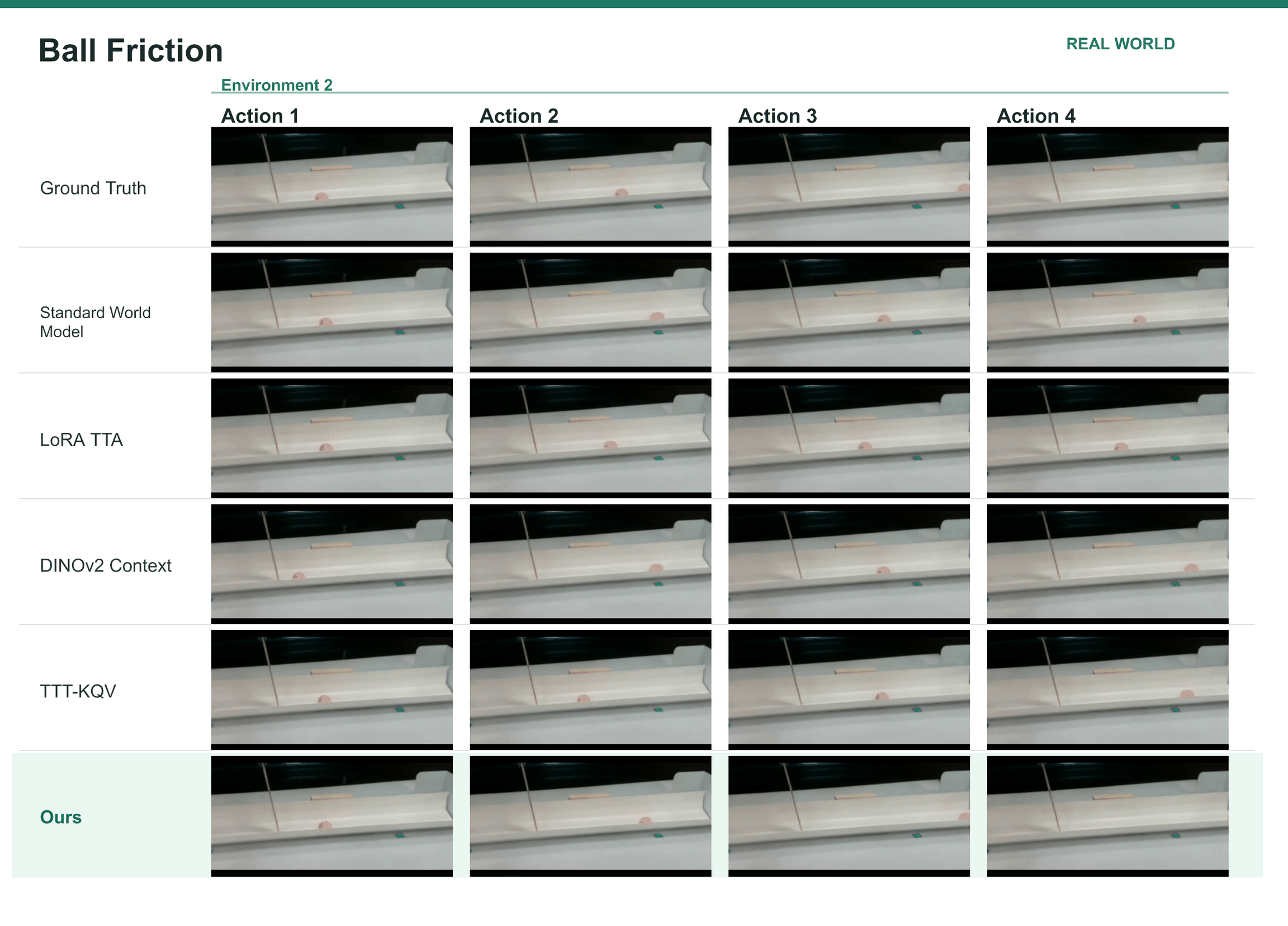}
\caption{Ball Friction: selected test actions.}\label{fig:qual-ball}
\end{figure}

\begin{figure}[p]
\centering\includegraphics[width=.90\linewidth]{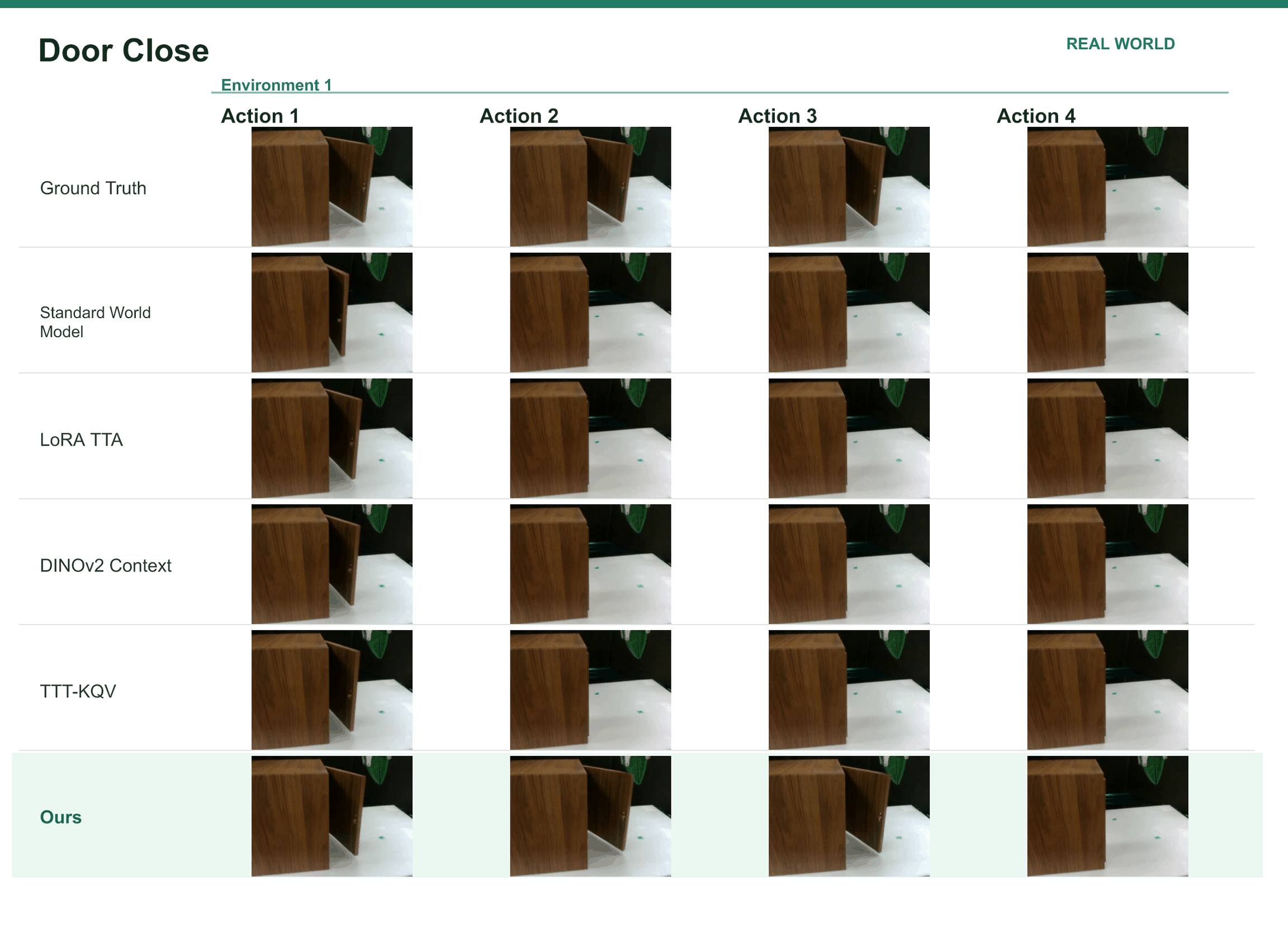}\par
\includegraphics[width=.90\linewidth]{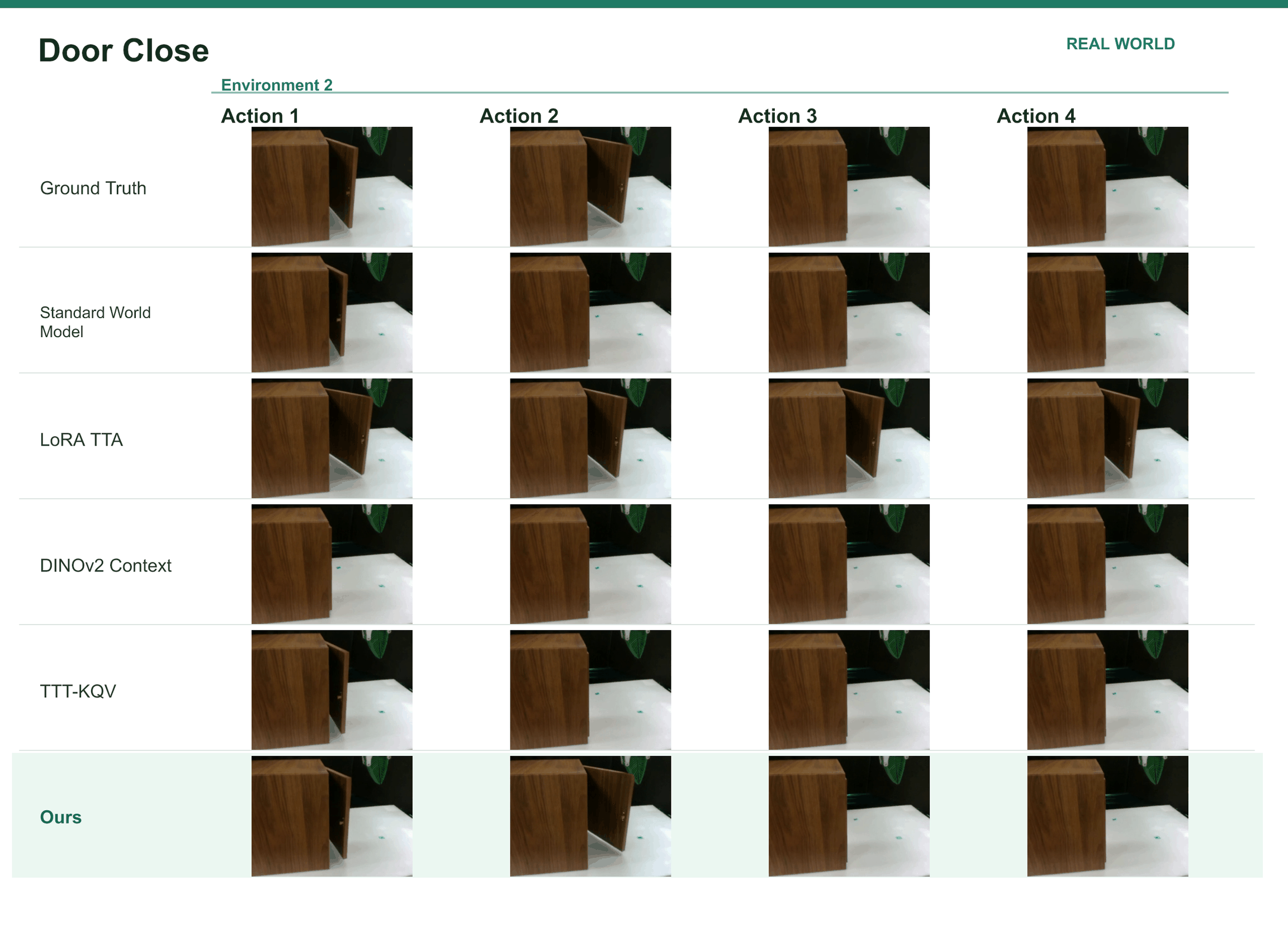}
\caption{Door Close: selected test actions.}\label{fig:qual-door}
\end{figure}

\begin{figure}[p]
\centering\includegraphics[width=.90\linewidth]{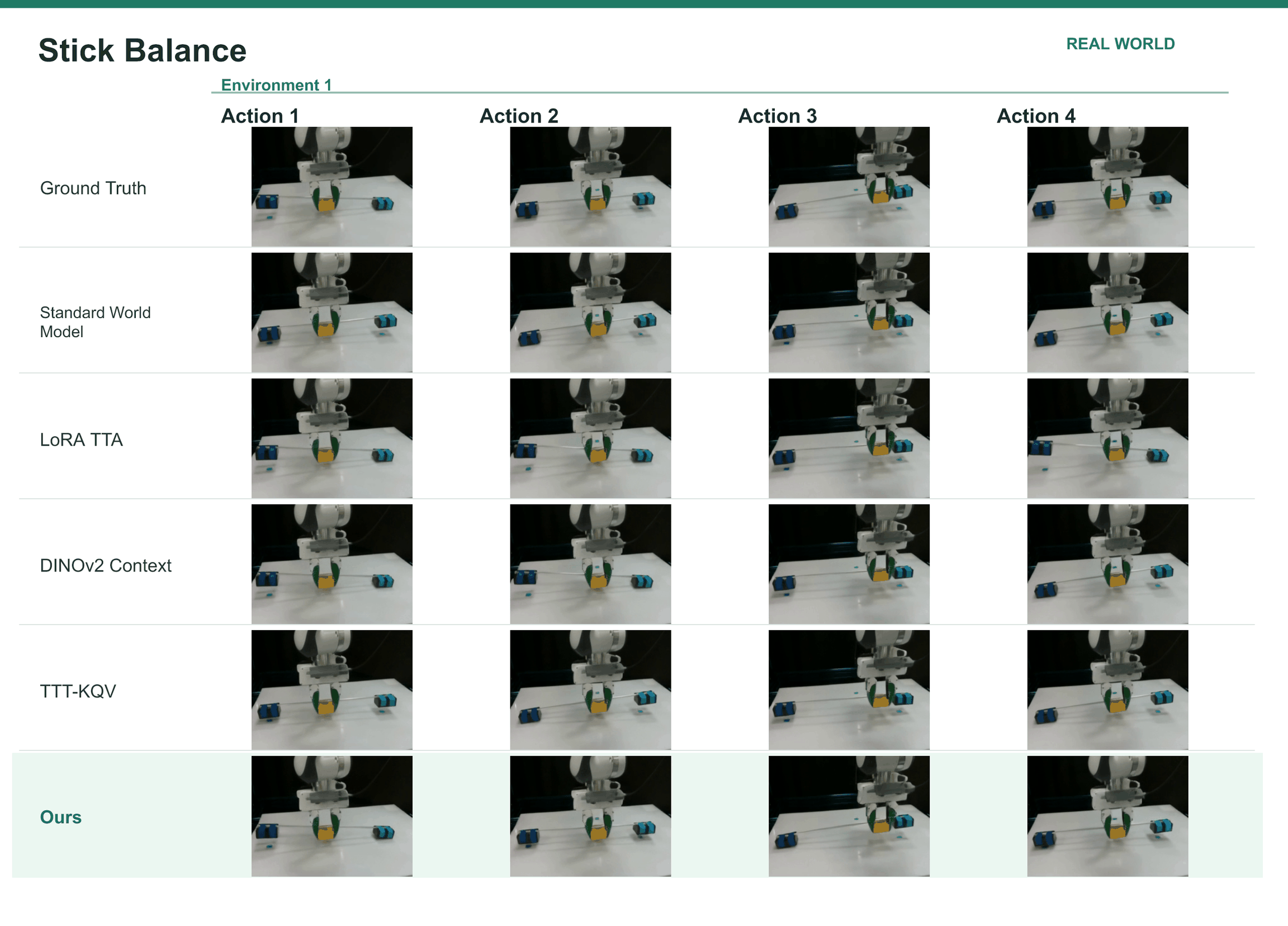}\par
\includegraphics[width=.90\linewidth]{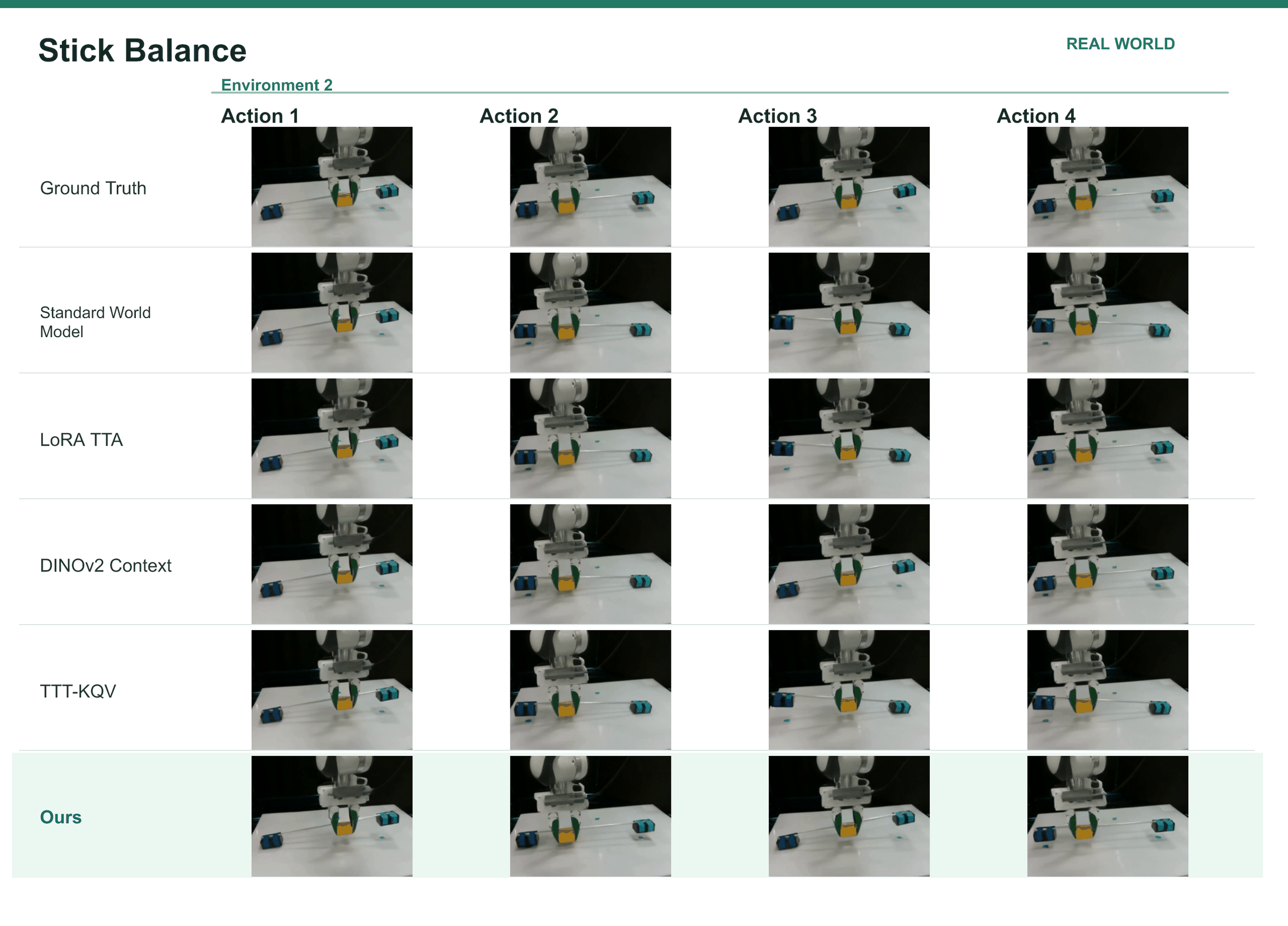}
\caption{Stick Balance: selected test actions.}\label{fig:qual-stick}
\end{figure}

\begin{figure}[!ht]
\centering\includegraphics[width=\linewidth]{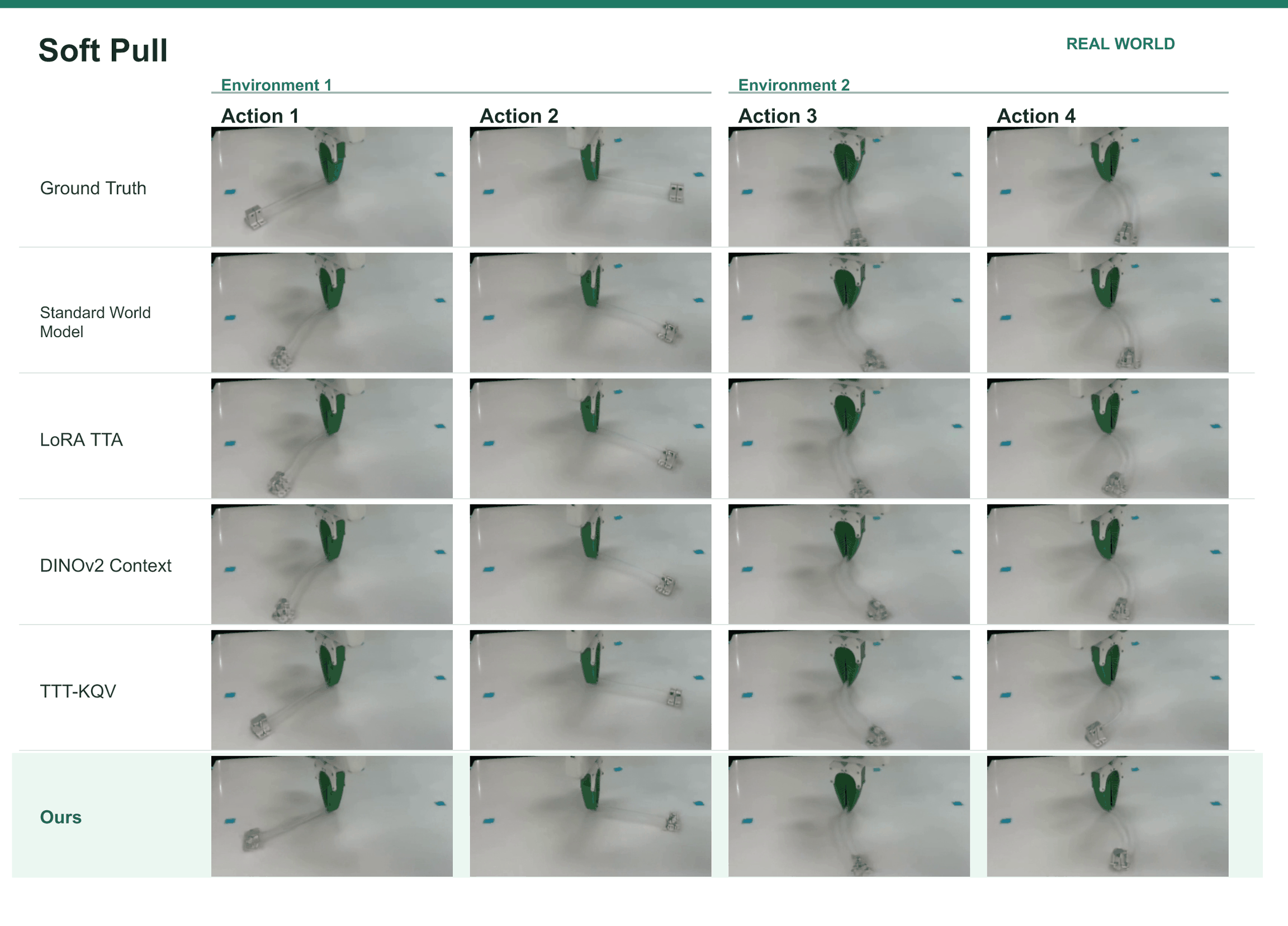}
\caption{Soft Pull: selected test actions.}\label{fig:qual-soft}
\end{figure}

\tzsection{Additional Qualitative Results}
\label{app:qualitative}
Figures~\ref{fig:qual-friction}--\ref{fig:qual-soft} show selected held-out
test actions. Each task shows two test environments and four actions per
environment; Soft Pull shows two actions per environment. Each column uses the
same initial observation and candidate
action across methods; rows show ground truth, the four baselines, and
our prediction. Simulation panels show the main camera only.

\clearpage

\ifdefined\SUPPSTANDALONE\else
\phantomsection

\fi

\end{document}